%% file: main.tex
\pdfoutput=1
\documentclass{article}
\usepackage[margin=1in]{geometry}
\usepackage{eqparbox}

\usepackage{microtype}
\usepackage{graphicx}
\usepackage{booktabs}

\usepackage{xcolor,colortbl}
\usepackage{microtype}      
\usepackage{multirow}
\usepackage{amsmath,amsfonts,amssymb, amsthm}
\usepackage{algorithm, algorithmic}
\usepackage{tikz}
\usepackage{pgfplots}
\usepackage{mathtools}
\usepackage{fixmath}
\usepackage{arydshln}
\usepackage{wrapfig}
\usepackage{enumitem}
\usepackage{graphicx}
\usepackage{transparent}
\usepackage{tabularx}
\usepackage{natbib}

\usepackage{float}

\usepackage{epstopdf}
\usepackage{caption}
\usepackage{subcaption}

\usepackage{caption}
\usepackage{calc}
\usepackage{relsize}
\usepackage{pifont}

\usepackage{tikz}

\usetikzlibrary{positioning, arrows.meta}

\usepackage{import}
\usepackage{pdfpages}

\usepackage[pdf]{graphviz}

\definecolor{myGreen}{HTML}{B6D7A8} 
\definecolor{color1}{HTML}{ef476f}
\definecolor{color2}{HTML}{f78c6b}
\definecolor{color3}{HTML}{ffd166}
\definecolor{color4}{HTML}{06d6a0}
\definecolor{color5}{HTML}{118ab2}
\definecolor{color6}{HTML}{073b4c}
\definecolor{color7}{HTML}{e377c2}

\definecolor{MidnightBlue}{RGB}{25, 25, 112}
\definecolor{Red}{rgb}{0.6,0,0}
\definecolor{lightgreen}{RGB}{171, 225, 175}

\newcommand{\firstmodality}{\mathbf{x}}
\newcommand{\secondmodality}{\mathbf{x'}}
\newcommand{\labely}{\mathbf{y}}

\usepackage[backref=page]{hyperref}
\hypersetup{colorlinks=true}
\hypersetup{linktoc=all}
\hypersetup{citecolor=MidnightBlue}
\hypersetup{linkcolor=Red}
\hypersetup{urlcolor=MidnightBlue}
\usepackage[all]{hypcap}
\usepackage[noabbrev, nameinlink, capitalize]{cleveref}
\input{math_commands.tex}

\newcommand{\pairpanel}[3]{%
  \begin{minipage}[c]{#1}
    \centering
    \begin{minipage}[c]{0.49\linewidth}
      \centering
      \includegraphics[width=\linewidth]{#2}
    \end{minipage}%
    \hspace{0.01\linewidth}%
    \begin{minipage}[c]{0.49\linewidth}
      \centering
      \includegraphics[width=\linewidth]{#3}
    \end{minipage}
  \end{minipage}%
}

\title{Characterizing the Predictive Impact of Modalities with \\ Supervised Latent-Variable Modeling}

\author{
    Divyam Madaan\thanks{New York University. Correspondence to \texttt{divyam.madaan@nyu.edu}}\\
  \and 
  Sumit Chopra\footnotemark[1] \textsuperscript{,}\thanks{New York University Grossman School of Medicine}
  \and 
  Kyunghyun Cho\footnotemark[1] \textsuperscript{,}\thanks{Genentech} \textsuperscript{,}\thanks{CIFAR} \\
}

\date{}
\pgfplotsset{compat=1.18}

\begin{document}
\maketitle

\begin{abstract}
Despite the recent success of Multimodal Large Language Models (MLLMs),
existing approaches predominantly assume the availability of 
multiple modalities during training and inference.
In practice, multimodal data is often incomplete because modalities may be missing, collected asynchronously, or available only for a subset of examples.
In this work, we propose \textbf{PRIMO}, a supervised latent-variable imputation model that quantifies the {\bf pr}edictive {\bf i}mpact of any missing {\bf mo}dality within the multimodal learning setting. PRIMO enables the use of all available training examples, whether modalities are complete or partial. Specifically, it models the missing modality through a latent variable that captures its relationship with the observed modality in the context of prediction. During inference, 
we draw many samples from the learned distribution over the missing modality to both obtain the marginal predictive distribution (for the purpose of prediction) and analyze the impact of the missing modalities on the prediction for each instance.
We evaluate PRIMO on a synthetic XOR dataset, Audio-Vision MNIST, and MIMIC-III for mortality and ICD-9 prediction. Across all datasets, PRIMO obtains performance comparable to unimodal baselines when a modality is fully missing and to multimodal baselines when all modalities are available. 
PRIMO quantifies the predictive impact of a modality at the instance level using a variance-based metric computed from predictions across latent completions. 
We visually demonstrate how varying completions of the missing modality result in a set of plausible labels.

\end{abstract}

\input{sections/1_introduction}

\input{sections/2_method}
\input{sections/3_related_work}
\input{sections/4_experiments}

\input{sections/5_conclusion}

\newpage
\bibliography{strings, references}
\bibliographystyle{tmlr}

\appendix
\input{sections/6_appendix}

\end{document}

%% file: math_commands.tex
\usepackage{amsmath,amsfonts,bm}

\newcommand{\observedmodality}{\mathbf{x}_{\text{o}}}
\newcommand{\missingmodality}{\mathbf{x}_{\text{m}}}
\newcommand{\latent}{\mathbf{z}}
\def\eqref#1{equation~\ref{#1}}
\def\1{\bm{1}}

\DeclareMathAlphabet{\mathsfit}{\encodingdefault}{\sfdefault}{m}{sl}
\SetMathAlphabet{\mathsfit}{bold}{\encodingdefault}{\sfdefault}{bx}{n}

%% file: sections/1_introduction.tex
\section{Introduction}
A central challenge in practical multimodal learning is the limited availability of all modalities for a downstream task. Many curated benchmarks, both in healthcare~\citep{soenksen_integrated_2022,huang2025hist,gu_illusion_2025} 
and in standard multimodal learning~\citep{antol_vqa_2015,goyal_making_2017,dancette_beyond_2021,tong_eyes_2024,liu_mmbench_2024, yue_mmmu_2024, wu_v_2024}, assume that all modalities are observed at training and inference. 

In practice, modalities are missing for many instances, especially in healthcare, where paired data is often incomplete~\citep{kleist_evaluation_2023, kleist_evaluation_2025, erion_cost-aware_2022}. When a patient arrives at the hospital, only a limited set of measurements may be collected initially, and additional tests are ordered only when clinicians suspect a specific condition. This matters because acquiring additional modalities can be expensive and can pose risks to patients. For instance, in prostate cancer screening, MRI before biopsy can improve downstream decision-making, but it also adds cost and exposes patients to additional procedures and potential risks~\citep{callender_benefit_2021}. 

In these settings, the goal is not to fill in the missing inputs, but to understand what the missing modality would actually change for the prediction. This motivates the central question of our work:
\begin{center}
\emph{For a given multimodal example, how does a modality affect the prediction?}
\end{center}
Most existing approaches model missing modalities as an imputation problem. They infer the missing modality conditioned on the observed modality and then treat the imputed value as observed. Many methods use generative models~\citep{suzuki_joint_2017, wu_multimodal_2018, shi_variational_2019, sutter_multimodal_2020, sutter_generalized_2021, joy_learning_2022, palumbo_mmvae_2023} 
to tackle this problem during inference by optimizing a variational lower bound on the data likelihood. This objective prioritizes reconstructing the input modalities; however, improved generative modeling does not necessarily translate into better discriminative performance.
This is because there can be many ways to fill in a modality, and only some of them matter for prediction. \citet{mancisidor_discriminative_2024} 
partially mitigates this issue by incorporating a discriminative objective, but assumes fully observed multimodal training data. Other approaches discard partially observed examples and either use only complete training data~\citep{suzuki_joint_2017, wu_multimodal_2018, shi_variational_2019, sutter_multimodal_2020, sutter_generalized_2021} or make predictions only on fully paired data~\citep{lee_multimodal_2023, wu_multimodal_2024}. None of these methods jointly optimize a discriminative objective while supporting partially observed modalities during both training and inference.

\begin{figure*}[t!]
    \centering
    \includegraphics[width=\linewidth]{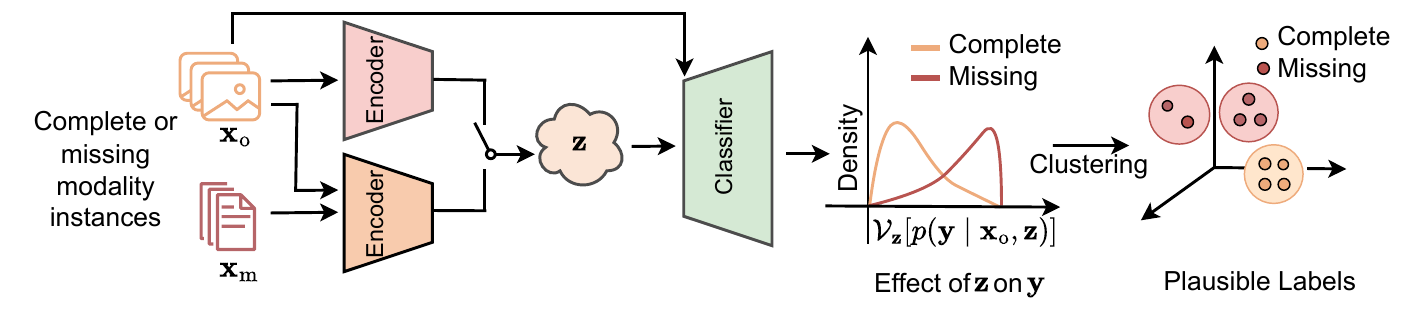}
    \vspace{-0.25in}
    \caption{{\bf Overview of PRIMO.}{ Given an observed modality $\observedmodality$ and an additional modality $\missingmodality$ that may be missing, PRIMO samples a latent variable $\latent$ conditioned on the available modalities. The classifier maps $(\observedmodality,\latent)$ to predictions, and the conditional variance $\mathcal{V}_{\latent}\left[p(\labely \mid \observedmodality,\latent)\right]$ quantifies how changes in $\latent$ affect the prediction. When both modalities are observed (orange), $\mathcal{V}$ is lower. When a modality is missing (red), $\mathcal{V}$ is higher. PRIMO then clusters the output logits across latent samples to visualize plausible labels under each availability scenario.}}
    \label{fig:concept_figure}
    \vspace{-0.1in}
\end{figure*}

We thus need an approach that (i) uses both complete and partially observed examples during training and inference, and (ii) captures uncertainty in the missing modality pertaining to the predictions for each instance. The goal is not to produce a single value of the missing modality, but to characterize how different plausible completions of the missing modality would change the predictive distribution for a given instance. To achieve this, we propose \textbf{PRIMO}, a supervised latent-variable model that quantifies the {\bf pr}edictive {\bf i}mpact of any {\bf mo}dality.
At a high level, PRIMO measures the impact of each missing modality by modeling it as a latent variable. 
During inference, PRIMO draws many samples from the learned distribution over the missing modality to obtain a set of final predictions that captures the marginal predictive distribution and the uncertainty due to the missing modality (see \Cref{fig:concept_figure}). 

More formally, let $\observedmodality$ denote the observed modality, $\missingmodality$ denote the additional modality that may be missing for some instances, and a target label $\labely$. 
Since modalities can be high-dimensional, directly modeling what part of $\missingmodality$ is relevant for $\labely$ can be challenging. We thus use a continuous latent variable $\latent$ to capture the information associated with the missing modality that is relevant for predicting $\labely$.
PRIMO is trained end-to-end to maximize the predictive distribution $p(\labely \mid \observedmodality)$ when $\missingmodality$ is unavailable and $p(\labely \mid \observedmodality, \missingmodality)$ when both modalities are observed. 
When $\missingmodality$ is absent during inference, $\latent$ is sampled from a conditional prior $p(\latent \mid \observedmodality)$. 
When both modalities are available, it is sampled from $p(\latent \mid \observedmodality, \missingmodality)$.

This latent-variable formulation enables the characterization of predictive impact due to the missing modality for each instance. Particularly, we measure
$\mathcal{V}_{\latent}\!\left[p(\labely \mid \observedmodality, \latent)\right]$ to quantify the effect of changes in $\latent$ on the output predictions. 
Small values of $\mathcal{V}$ imply that the output is less dependent on the missing modality, whereas large values indicate a greater dependence.
The distribution over logits yields instance-level estimates of modality impact and captures the range of plausible predictions induced by different latent samples. This also allows us to use PRIMO as a diagnostic tool in complete-modality scenarios to test modality dependence and identify when multimodal models rely on shortcuts~\citep{fu_blink_2024,tong_eyes_2024,madaan_multi-modal_2025, gu_illusion_2025}.

We evaluate PRIMO on synthetic and real-world multimodal benchmarks. 
These include a synthetic XOR dataset, Audio-Vision MNIST~\citep{liang_multibench_2021} 
with audio and vision modalities, and MIMIC-III~\citep{johnson_mimic-iii_2016, liang_multibench_2021} with patient demographics and clinical time-series for mortality and ICD-9 code prediction.
Across all datasets, PRIMO obtains performance comparable to unimodal baseline $p(\labely \mid \observedmodality)$ 
when a modality is missing, and to a multimodal baseline $p(\labely \mid \observedmodality, \missingmodality)$ when all modalities are available. Beyond predictive performance, PRIMO provides insight into the impact of different modalities for a given task. For example, we show that patient demographic information is sufficient for mortality prediction and neoplasm ICD-9 code prediction in MIMIC-III, while clinical time-series is essential for respiratory ICD-9 code prediction.

%% file: sections/2_method.tex
\section{Learning with Both Complete and Missing Modalities}

{We consider supervised multimodal learning where a modality can be missing during training and inference. 
For clarity, we focus on two modalities. Each example consists of an observed modality 
$\observedmodality$, an additional modality $\missingmodality$ that may be absent, 
and a label $\labely\in \Delta^{C-1}$ over $C$ classes. 
The dataset contains complete examples $\mathcal{D}_{\text{complete}}=\{(\mathbf{x}_{\text{o},i},\mathbf{x}_{\text{m},i},\mathbf{y}_i)\}_{i=1}^{N_c}$ 
and missing-modality examples $\mathcal{D}_{\text{missing}}=\{(\mathbf{x}_{\text{o},j},\mathbf{y}_j)\}_{j=1}^{N_m}$.
PRIMO learns a predictor that maps the available modalities to $\labely$, 
using $(\observedmodality, \missingmodality)$ when $\missingmodality$ is present, and only $\observedmodality$ otherwise.}

To characterize the impact of a missing modality, our goal is not to reconstruct $\missingmodality$ but to capture the uncertainity in $\missingmodality$ that is relevant for prediction. PRIMO is a supervised latent-variable model trained end-to-end (Section~\ref{sec:training}) that supports both complete and missing-modality inputs. It samples latent completions from $p(\latent \mid \observedmodality)$ when $\missingmodality$ is missing and from $p(\latent \mid \observedmodality,\missingmodality)$ when both modalities are observed, which typically reduces predictive variance. This enables us to quantify how predictions vary across completions and use PRIMO for both prediction and modality impact analysis at inference time (Section~\ref{sec:inference}).

\subsection{Learning objective}\label{sec:training}
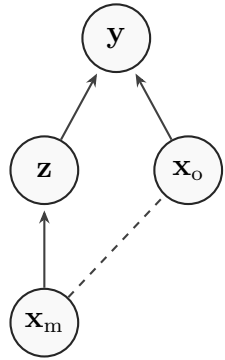
\begin{wrapfigure}{r}{0.27\textwidth}
  \centering
  \vspace{-25pt}
  \begin{tikzpicture}[
    node distance=1.1cm and 0.6cm,
    >={Stealth[length=5pt, width=4pt]}
  ]
    \node[circle, draw=black!85, fill=gray!5, thick, minimum size=0.9cm, font=\large] (y) {$\labely$};
    \node[circle, draw=black!85, fill=gray!5, thick, minimum size=0.9cm, font=\large] (z) [below left=of y, xshift=0.3cm] {$\latent$};
    \node[circle, draw=black!85, fill=gray!5, thick, minimum size=0.9cm, font=\large] (xo) [below right=of y, xshift=-0.3cm] {$\observedmodality$};
    \node[circle, draw=black!85, fill=gray!5, thick, minimum size=0.9cm, font=\large] (xm) [below=of z] {$\missingmodality$};

    \draw[->, thick, color=black!75, shorten >=2pt] (z) -- (y);
    \draw[->, thick, color=black!75, shorten >=2pt] (xo) -- (y);
    \draw[->, thick, color=black!75, shorten >=2pt] (xm) -- (z);
    
    \draw[dashed, thick, color=black!75] (xm) -- (xo);
  \end{tikzpicture}
  \caption{DGP for missing modalities. {The dashed line denotes an a priori correlation between the two modalities.}\label{fig:dgp}}
  \vspace{-0.6in}
\end{wrapfigure}

We optimize variational lower bounds on the conditional log-likelihoods
$\log p(\labely \mid \observedmodality, \missingmodality)$ and
$\log p(\labely \mid \observedmodality)$.
Following the data generating process (DGP) in \Cref{fig:dgp}, we model the label-relevant information 
in the missing modality $\missingmodality$ with a continuous latent variable 
$\latent\in~\mathbb{R}^d$. 
We assume that $\labely$ is conditionally independent of $\missingmodality$ given $(\observedmodality, \latent)$. Under this assumption, the predictive distributions for complete and missing-modality inputs are
\begin{equation}\label{eq:pred_missing_modality}
\begin{aligned}
p(\labely\mid \observedmodality, \missingmodality)
&= \int p_\theta(\labely \mid \observedmodality,\latent)\,
p_\omega(\latent \mid \observedmodality,\missingmodality)\,d\latent, \\
p(\labely\mid \observedmodality)
&= \int p_\theta(\labely\mid \observedmodality, \latent)\,
p_\omega(\latent\mid \observedmodality)\,d\latent, 
\end{aligned}
\end{equation}
where $p_\theta$ is the predictive model and $p_\omega$ parameterizes the conditional latent distributions.
Since these integrals are intractable, we introduce an approximate posterior $q_\phi$ and maximize the resulting evidence lower bounds (ELBOs) for both scenarios (see \Cref{sec:elboderivation} of the Appendix for complete derivations).

\paragraph{Case 1: Complete modalities} ($\mathcal{D}_{\text{complete}}$).
When both modalities are observed, we approximate the true posterior
$p(\latent \mid \observedmodality, \missingmodality, \labely)$ with
$q_\phi(\latent \mid \observedmodality, \missingmodality, \labely)$ as the variational posterior.
We use a conditional prior $p_\omega(\latent \mid \observedmodality, \missingmodality)$.
We maximize the following ELBO:
\begin{equation}
\label{eq:elbo_complete}
\begin{aligned}
\mathcal{L}_{\mathrm{complete}}^{\mathrm{ELBO}} 
=
\mathbb{E}_{\latent \sim q_\phi(\latent \mid \observedmodality, \missingmodality, \labely)}
\Big[
    \log p_\theta(\labely \mid \observedmodality, \latent)
\Big] 
- \operatorname{KL}\!\left(
    q_\phi(\latent \mid \observedmodality, \missingmodality, \labely)
    \,\Vert\,
    p_\omega(\latent \mid \observedmodality, \missingmodality)
\right).
\end{aligned}
\end{equation}

\paragraph{Case 2: Missing modality} ($\mathcal{D}_{\text{missing}}$).
When $\missingmodality$ is missing, we use the variational posterior
$q_\phi(\latent \mid \observedmodality, \labely)$ and the conditional prior
$p_\omega(\latent \mid \observedmodality)$.
The resulting ELBO is
\begin{equation}
\label{eq:elbo-missing}
\begin{aligned}
\mathcal{L}_{\mathrm{missing}}^{\mathrm{ELBO}} 
= \mathbb{E}_{\latent \sim q_\phi(\latent \mid \observedmodality, \labely)}
\Big[
    \log p_\theta(\labely \mid \observedmodality, \latent)
\Big]
- \operatorname{KL}\!\left(
    q_\phi(\latent \mid \observedmodality, \labely)
    \,\Vert\,
    p_\omega(\latent \mid \observedmodality)
\right).
\end{aligned}
\end{equation}
We jointly maximize both ELBOs across the training set to learn a shared latent representation in complete and missing-modality scenarios. Both ELBOs maximize $\log p_\theta(\labely \mid \observedmodality, \latent)$ and contain no reconstruction term for the missing modality.

{When trained jointly, the unimodal and multimodal conditional priors can shift together in $\latent$ without changing the KL. Because KL divergence is translation invariant, a common shift of both distributions leaves the KL invariant, creating a shift symmetry in $\latent$. We break this symmetry by anchoring $p_\omega(\latent\mid\observedmodality)$ to $\mathcal{N}(\mathbf{0},\mathbf{I})$~\citep{mansimov2019molecular} and tying $p_\omega(\latent\mid\observedmodality,\missingmodality)$ to $p_\omega(\latent\mid\observedmodality)$ for the same $\observedmodality$ using a regularizer $\mathcal{R}$:}
\begin{equation} 
\label{eq:regularizer} 
\begin{aligned} 
\mathcal{R} = \sum_{i=1}^{N_c+N_m} \operatorname{KL}\!\left(p_\omega(\latent \mid \mathbf{x}_{\text{o},i}) \,\Vert\, \mathcal{N}(\mathbf{0},\mathbf{I}) \right) 
+ \sum_{i=1}^{N_c} \operatorname{KL}\!\left( p_\omega(\latent \mid \mathbf{x}_{\text{o},i}, \mathbf{x}_{\text{m},i}) \,\Vert\, p_\omega(\latent \mid \mathbf{x}_{\text{o},i}) \right), 
\end{aligned}
\end{equation}
We parameterize the conditional priors $p_\omega(\latent \mid \cdot)$ and the variational posteriors
$q_\phi(\latent \mid \cdot)$ as diagonal Gaussians, where the mean and variance are given by shared amortized
networks. The conditioning variables depend on modality availability, with the priors conditioned on $\observedmodality$ or $(\observedmodality,\missingmodality)$, and the posteriors conditioned on $(\observedmodality,\labely)$ or $(\observedmodality,\missingmodality,\labely)$.
\begin{equation}
\begin{aligned}
p_\omega(\latent\mid \cdot)
=
\mathcal{N}\!\left(
\latent;\,
\mu_\omega(\cdot),\,
\operatorname{diag}\!\big(\sigma_\omega(\cdot)^2\big)
\right), \quad
q_\phi(\latent\mid \cdot)
=
\mathcal{N}\!\left(
\latent;\,
\mu_\phi(\cdot),\,
\operatorname{diag}\!\big(\sigma_\phi(\cdot)^2\big)
\right).
\end{aligned}
\end{equation}
To prevent posterior collapse, we follow \citet{zhu_batch_2020} and apply batch normalization
($\operatorname{BN}$) to the posterior mean $\mu_\phi(\cdot)$ with fixed scale $\gamma$ and learnable offset $\beta$.
We compute BN statistics using mini-batch statistics during training. This prevents the posterior from trivially matching the prior by encouraging the KL term to remain non-zero.
During training, we use the reparameterization trick to allow backpropagation through samples from these
distributions:
\begin{equation}
\latent
=
\mu_\phi(\cdot)
+
\sigma_\phi(\cdot)\odot\boldsymbol{\varepsilon},
\qquad
\boldsymbol{\varepsilon}\sim\mathcal{N}(\mathbf{0},\mathbf{I}),
\end{equation}
where $\odot$ denotes the element-wise (Hadamard) product. The final training objective is
\begin{equation}
\max_{\theta,\phi,\omega}
\;
\sum_{i=1}^{N_c}
\mathcal{L}^{\mathrm{ELBO}}_{\mathrm{complete},i}
+
\sum_{j=1}^{N_m}
\mathcal{L}^{\mathrm{ELBO}}_{\mathrm{missing},j}
-
\mathcal{R}, 
\label{eq:final_objective}
\end{equation}
where we optimize jointly over all model parameters.

\subsection{Inference}\label{sec:inference}
During testing, the labels $\labely$ are unknown.
We obtain predictions by marginalizing out the latent variable under the appropriate conditional prior and approximate the resulting integral via Monte Carlo sampling. We draw $K$ latent samples from the prior
and average the resulting predictive probabilities:
\begin{equation}
\label{eq:test_pred}
\begin{aligned}
p_\theta(\labely\mid \observedmodality)
&\approx
\frac{1}{K}\sum_{k=1}^K
p_\theta(\labely\mid \observedmodality,\latent^{(k)}), \quad
\latent^{(k)}&\sim p_\omega(\latent\mid \observedmodality).
\end{aligned}
\end{equation}
{Following \Cref{fig:dgp}, when both modalities are available at test time, we use the complete prior $p_\omega(\latent \mid \observedmodality, \missingmodality)$.}

To evaluate whether the missing modality is informative, we measure how the predictions change as we vary latent samples. 
We define $\mathcal{V}\equiv \mathcal{V}_{\latent}\!\left[p_\theta(\cdot \mid \observedmodality,\latent)\right]$ as the expected total variation distance (TVD) between the predictive distribution
$p_\theta(\cdot \mid \observedmodality,\latent)$ and its mean given by
$\bar{p}_\theta(\cdot \mid \observedmodality)=\mathbb{E}_{\latent \sim p_\omega(\latent \mid \observedmodality)}[p_\theta(\cdot \mid \observedmodality, \latent)]$: 
\begin{equation}
\label{eq:variance}
\begin{aligned}
\mathcal{V}
=
\mathbb{E}_{\latent \sim p_\omega(\latent \mid \observedmodality)}
\Big[
\mathrm{TVD}\!\left(
p_\theta(\cdot \mid \observedmodality, \latent),
\bar{p}_\theta(\cdot \mid \observedmodality)
\right)
\Big].
\end{aligned}
\end{equation}
We denote this quantity by $\mathcal{V}_{\text{missing}}$ when $\latent \sim p_\omega(\latent \mid \observedmodality)$, and by $\mathcal{V}_{\text{complete}}$ when $\latent \sim p_\omega(\latent \mid \observedmodality,\missingmodality)$.
Larger $\mathcal{V}_{\text{missing}}$ indicates that $\missingmodality$ can substantially alter the predictions.

To characterize plausible outputs under missingness, we draw $\latent \sim p_\omega(\latent \mid \observedmodality)$, get the corresponding output logits from $p_\theta(\cdot \mid \observedmodality,\latent)$, and cluster these logits using a Dirichlet Process Gaussian Mixture Model (DPGMM). We label each cluster by its mean predicted class distribution, yielding a set of plausible labels for the input. If the clusters contain multiple plausible labels, this indicates that the latent variable (and thus the missing modality) significantly influences the prediction. Conversely, if the clusters are dominated by a single label, this suggests that the observed modality is sufficient.

%% file: sections/3_related_work.tex
\section{Related Work}

\paragraph{Data imputation.}
Missing data imputation has been studied extensively outside multimodal learning. Earlier works used simple heuristics such as zero-filling~\citep{liu_m3ae_2023, parthasarathy_training_2020} and averaging-based variants such as mean/mode imputation or nearest neighbors. 
Many benchmarks evaluate imputation methods under different datasets and missingness assumptions~\citep{luengo_choice_2012, poulos_missing_2018, woznica_does_2020, le_morvan_whatsa_2021, shadbahr_impact_2023, li_comparison_2024, morvan_imputation_2025}. These evaluations focus on imputation quality rather than downstream predictive performance. 
Prior works have shown that improved imputations do not always translate to better downstream accuracy~\citep{shadbahr_impact_2023, morvan_imputation_2025}.
Under the missing completely at random assumption, \citet{paterakis_we_2024} similarly reports limited gains beyond simple mean and mode baselines. 
Similarly to our work, \citet{ramchandran_learning_2024} proposes a latent-variable model that treats missing covariates as latent variables and marginalizes them out during inference. Their model, however, differs from ours in that it creates a single latent variable for all covariates, and their analysis does not investigate how the imputation distribution affects the predictive distribution in a fine-grained manner. 

\paragraph{Multimodal learning with missing modalities.}
Many Variational Autoencoder (VAE)-based multimodal models have also been proposed to handle missing modalities~\citep{suzuki_joint_2017,vedantam_generative_2018, tsai_learning_2019, shi_variational_2019,sutter_generalized_2021,gong_variational_2021,joy_learning_2022, palumbo_mmvae_2023}. These methods focus on generative modeling by optimizing a marginal-likelihood objective via an ELBO, learning to reconstruct the inputs while regularizing the latent distribution toward a prior.
As a result, $\latent$ captures variation in the inputs, but it does not align $\latent$ with the discriminative decision boundary required for modeling $p(\labely\mid \cdot)$ under missing modalities. 
CMMD~\citep{mancisidor_discriminative_2024} takes a step in this direction by incorporating a discriminative component into the multimodal latent framework, but assumes fully observed data during training.  
MEME~\citep{joy_learning_2022} and VSVAE~\citep{gong_variational_2021} consider
partial modality availability, where only a subset of training examples contains
all modalities, but they also focus on generative modeling.  
In contrast, PRIMO focuses on discriminative prediction under heterogeneous modality availability during both training and inference. Additional details of these methods are provided in \Cref{appendix:relatedwork} of the Appendix.

\paragraph{Multimodal learning with complete modalities.}
One aspect of multimodal learning that has gained interest in recent years is the propensity of multimodal models to rely on a single modality rather than utilizing all available modalities~\citep{agrawal_dont_2018, singh_towards_2019, dancette_beyond_2021, si_check_2021, madaan_jointly_2024, yue_mmmu-pro_2025}. 
More recently, the community has thus shifted its attention to analyzing these multimodal datasets by using various diagnostic checks and metrics. 
These include measuring performance change under modality removal or shuffling~\citep{gu_illusion_2025, madaan_multi-modal_2025}, defining modality importance scores~\citep{gat_perceptual_2021, park_assessing_2024}, or circular evaluation~\citep{liu_mmbench_2024}. 
These approaches often lack either a way to analyze these multimodal data at an instance level or a mathematically interpretable justification. Our approach, PRIMO, on the other hand, allows us to inspect the impact of a (missing) modality at the level of individual instances in a fine-grained manner.

%% file: sections/4_experiments.tex
\input{figures/figure_1}

\input{figures/figure_2}

\input{figures/figure_3}

\section{Experiments}
We evaluate the effectiveness of PRIMO on a diverse set of multimodal datasets spanning synthetic, vision-audio, and healthcare settings.
We use a synthetic XOR dataset, Audio-Vision MNIST~\citep{liang_multibench_2021} with missing audio or vision, and MIMIC-III~\citep{johnson_mimic-iii_2016, liang_multibench_2021} with patient demographics (static) and clinical measurements (time-series). Additional dataset, hyperparameters, and architecture details are provided in \Cref{appendixsection:experiments} of the Appendix.

Across all datasets, we compare PRIMO under both complete and missing modality conditions against (i) a unimodal baseline that observes only $\observedmodality$, and (ii) a multimodal $(\observedmodality, \missingmodality)$ baseline that observes both modalities when available. 
To evaluate when a missing modality is informative, we use our proposed metric $\mathcal{V}$ and the clustering analysis defined in \Cref{sec:inference}. 
We report ECDFs of $\mathcal{V}$ over the test set to summarize its instance-level distribution.

\subsection{Synthetic XOR}
{We consider two 1D modalities $\observedmodality$ and $\missingmodality$, always observing $\observedmodality$ and masking $\missingmodality$ at random with probability $0.5$. We sample $(\observedmodality,\missingmodality)$ from a mixture of three Gaussians with $\sigma=0.5$ centered at $(-1,-1)$, $(-1,1)$, and $(1,-1)$, and assign XOR labels based on the signs of $(\observedmodality,\missingmodality)$. This yields examples where for $\observedmodality<0$ the label depends on $\missingmodality$, while for $\observedmodality>0$ it can be determine by $\observedmodality$ only.}

\paragraph{Results.} {\Cref{fig:xor_results} (left) shows accuracy in complete and missing scenarios. Alongside unimodal and multimodal baselines, we compare against MVAE~\citep{wu_multimodal_2018} and MMVAE~\citep{shi_variational_2019} (generative baselines), CMMD~\cite{mancisidor_discriminative_2024} (discriminative missing-modality baseline), and LVAE~\citep{ramchandran_learning_2024} (imputation for missing covariates).}

{With $\missingmodality$ missing, all methods perform comparably to the unimodal baseline using only $\observedmodality$.
With complete inputs, only PRIMO and LVAE match the multimodal model evaluated on complete inputs, consistent with MVAE/MMVAE not being optimized for classification. In our setup, CMMD is not directly applicable to the complete-modality scenario because during inference it always uses the conditional prior $p_\omega(\latent \mid \observedmodality)$, even when $\missingmodality$ is observed.}

\Cref{fig:xor_results} (right) shows the predictive impact gap between missing and complete scenarios, $\mathcal{V}_{\text{missing}} - \mathcal{V}_{\text{complete}}$. Examples on the left exhibit a larger gap because the label depends on both modalities, whereas examples on the right are predictable from $\observedmodality$ alone. This demonstrates that PRIMO captures the predictive impact of the missing modality. \Cref{appendixsection:experiments} in the Appendix further visualizes the latent space and predictions across methods.

\subsection{Audio-vision MNIST (AV-MNIST)}

AV-MNIST is a multimodal digit classification dataset with ten classes using written digits from the MNIST dataset~\citep{lecun_gradient-based_1998} and human spoken digits from the Free Spoken Digit Dataset (FSDD)~\citep{jackson_jakobovskifree-spoken-digit-dataset_2018}. 
To control the task difficulty, the dataset variant introduced by \citet{liang_multibench_2021} varies the information content in each modality. For audio samples, it adds real-world environmental sounds from the ESC-50 dataset~\citep{piczak_esc_2015}. These are randomly selected from one of the ESC-50 categories. For image samples, it uses PCA-based energy reduction. 
We consider two missing-modality settings, masking either the audio modality or the image modality independently with probability $0.5$.

\paragraph{Results.} 
\Cref{tab:avmnist_acc_results} shows the accuracy when audio or vision modality is missing.
In both scenarios, PRIMO performs comparably to the unimodal ($\observedmodality$) and the multimodal I2M2~\citep{madaan_jointly_2024} baselines. 

We compare the distribution of $\mathcal{V}$ in both missing-modality scenarios in \Cref{fig:audio_missing_overlap} and \Cref{fig:image_missing_overlap}. 
Missing vision results in a significantly higher $\mathcal{V}$ ($\mu_{\text{miss}} = 0.57$) than missing audio ($\mu_{\text{miss}} = 0.37$). 
% This difference suggests that changing vision leads to a larger change in the predictions. 
In contrast, when audio is missing, many examples exhibit $\mathcal{V}$ comparable to the complete input setting, suggesting that the prediction is often insensitive to the audio modality for those instances.

In \Cref{fig:avmnistvisual}, we further characterize how the latent $\latent$ capturing the missing modality affects predictive distribution using the clustering analysis from \Cref{sec:inference}. We visualize which labels are most likely under the missing-audio (top row) and missing-vision (bottom row) settings. 
{We visualize high- and low-variance examples with their corresponding $\mathcal{V}_{\text{missing}}$ reported as $\mathcal{V}$. 
We observe that high-$\mathcal{V}$ examples often correspond to multiple plausible labels, reflecting that different latent completions of the missing modality can change the predicted label distribution.
For low-$\mathcal{V}$ examples, there is a concentration on a single dominant label across different latent completions in both settings.
These results illustrate that PRIMO 
captures how missing modalities alter the set of plausible predictions differently for different examples. 
}

\subsection{MIMIC-III}
MIMIC-III~\citep{johnson_mimic-iii_2016} is a clinical dataset that contains Electronic Health Records (EHR) data from approximately $40{,}000$ patients at Beth Israel Deaconess
Medical Center from 2001 to 2012. 
We use two modalities: (a) static modality containing patient-level information such as age, admission type, and chronic conditions (acquired immunodeficiency syndrome, hematologic malignancy, and metastatic cancer), and (b) time-series modality with 12 physiological measurements recorded hourly over the first 24 hours after ICU admission. 
The raw time-series modality contains missing measurements. We use the processed benchmark version~\citep{purushotham_benchmarking_2018,liang_multibench_2021}, which applies forward and backward filling, with mean imputation for features that are entirely missing. We investigate the predictive impact of the time-series modality across multiple tasks by masking it at random with probability $0.5$.
We consider mortality prediction as a 6-class problem (death within 1 day, 2 days, 3 days, 1 week, 1 year, or $>$1 year) and two {ICD-9 groups binary prediction tasks, where we consider Group 1 (codes 140–239, neoplasms) and Group 7 (codes 460–519, respiratory diseases)}.

\paragraph{Mortality prediction.} Time-series modality is often assumed to be important~\citep{liang_multibench_2021, madaan_jointly_2024} for this task since patient trajectories can capture deterioration patterns beyond static features. 
\Cref{tab:mimic_results} shows that the aggregate performance gain from including the time-series modality is relatively small. 
$\mathcal{V}$ in \Cref{fig:mimic_var} shows that for most patients, the predictions are stable across plausible completions of the time series modality.

{To investigate the tail of patients in this distribution, we conduct the clustering-based analysis from \Cref{sec:inference}. \Cref{fig:mortality_var} (left) shows that the resulting label distribution from the clusters shifts towards high-risk mortality outcomes as age increases. 
This suggests that time-series might be more informative for older-aged patients. 
We show plausible labels for individual patients in \Cref{fig:mimic_ind_patients} (in the main text) and \Cref{fig:mortality_results_appendix} (in the Appendix). 
Consistent with our results, we observe that low-$\mathcal{V}$ cases usually correspond to low-risk predictions. 
In contrast, high-$\mathcal{V}$ cases concentrate among patients closer to high-risk mortality classes, where time-series modality can be critical.}

\paragraph{ICD-9 code prediction.}
For predicting neoplasms (ICD-9 140--239), we obtain high accuracy in \Cref{tab:mimic_results} and low $\mathcal{V}_{\text{missing}}$ in \Cref{fig:mimic_var} despite the absence of time-series modality. 
This suggests that static modality is sufficient for this task.
This is consistent with our clustering analysis in \Cref{fig:mimic_ind_patients} and \Cref{fig:appendix_icd1_results}, where the predictions do not change much even in high-$\mathcal{V}$ samples and are dominated by a single label. This is supported by static modality containing chronic disease features, which are informative descriptors for this ICD block (see \Cref{fig:mortality_var} (right)).

In contrast, for respiratory disease (ICD-9 460-519), we obtain near-random performance in \Cref{tab:mimic_results} and high $\mathcal{V}_{\text{missing}}$ when the time-series modality is missing in \Cref{fig:mimic_var}. This is because respiratory diagnoses in the ICU depends on various time-series measurements. 
For instance, oxygenation-related variables such as PaO$_2$/FiO$_2$ are direct indicators of respiratory impairment, while other features such as temperature, WBC count, heart rate capture systemic instability, and infection that often co-occur with respiratory disease. 
Our patient-level analysis in \Cref{fig:mimic_ind_patients} and \Cref{fig:icd2_results} shows that missing time series leads to high-$\mathcal{V}_{\text{missing}}$ and ambiguity in the output predictions for most examples. 

Overall, for a given dataset, depending on the task, modality importance can vary significantly. 
While for neoplasms, time series is often not essential, it plays an important role for respiratory diseases. 
These results show that PRIMO captures these nuances, obtaining good performance while providing patient-level analysis.

\subsection{Bias Analysis}

Let $p^*(\cdot \mid \observedmodality)=\mathbb{E}[\labely \mid \observedmodality]$ denote the Bayes-optimal predictor given only $\observedmodality$. PRIMO defines a predictive distribution conditioned on $(\observedmodality,\latent)$, $p_\theta(\cdot \mid \observedmodality,\latent)$. Marginalizing over the learned conditional prior gives
\begin{equation}
\bar{p}_\theta(\cdot \mid \observedmodality)
\;=\;
\mathbb{E}_{\latent \sim p_\omega(\latent \mid \observedmodality)}
\big[p_\theta(\cdot \mid \observedmodality, \latent)\big].
\end{equation}
We measure the discrepancy between this mean prediction and the Bayes-optimal unimodal predictor using:
\begin{align}
\label{eq:bv-final}
\mathcal{B}_{\text{missing}}
= \mathrm{TVD}\!\left(
p^*(\cdot \mid \observedmodality),\;
\bar{p}_\theta(\cdot \mid \observedmodality)
\right).
\end{align}
When both modalities $(\observedmodality,\missingmodality)$ are observed, we define $\bar{p}_\theta(\cdot \mid \observedmodality,\missingmodality)$ analogously by replacing the prior with $p_\omega(\latent \mid \observedmodality,\missingmodality)$, and we define $\mathcal{B}_{\text{complete}}$ by comparing to the Bayes-optimal multimodal predictor $p^*(\cdot \mid \observedmodality,\missingmodality)=\mathbb{E}[\labely \mid \observedmodality,\missingmodality]$. This quantifies how well the learned priors recover the Bayes-optimal unimodal and multimodal predictors after marginalizing over $\latent$.

\begin{wrapfigure}{r}{0.47\textwidth}
    \centering
    \vspace{-0.2in}
    \includegraphics[width=\linewidth]{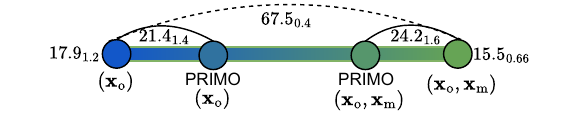}
    \vspace{-0.35in}
    \caption{{\bf Bias analysis with vision missing.} PRIMO$(\mathbf{x}_{\mathrm{o}})$ stays close to the unimodal oracle, while PRIMO$(\mathbf{x}_{\mathrm{o}},\mathbf{x}_{\mathrm{m}})$ stays close to the multimodal oracle. The dashed arc shows the unimodal--multimodal oracle gap.}
    \label{tab:bias_analysis}
\end{wrapfigure}

To obtain an unbiased estimate, we partition the dataset into two disjoint halves. Using the complete-modality half, we train unimodal and multimodal oracles. Using the remaining half, we train PRIMO with a $50\%$ missing rate and evaluate it under the same missingness pattern at inference time for both missing and complete scenarios.

\Cref{tab:bias_analysis} shows the bias for the missing-vision setting. The oracle distances provide a practical lower bound and are non-zero due to finite-sample effects and optimization noise. Under missing vision modality, PRIMO ($\observedmodality$) is closer to the unimodal oracle trained on $\observedmodality$. When both modalities are available, PRIMO ($\observedmodality,\missingmodality$) is closer to the multimodal oracle trained on $(\observedmodality,\missingmodality)$, consistent with the unimodal--multimodal oracle gap induced by observing $\missingmodality$.

%% file: figures/figure_1.tex
\begin{figure*}[htbp]
    \centering
    \includegraphics[width=0.54\linewidth]{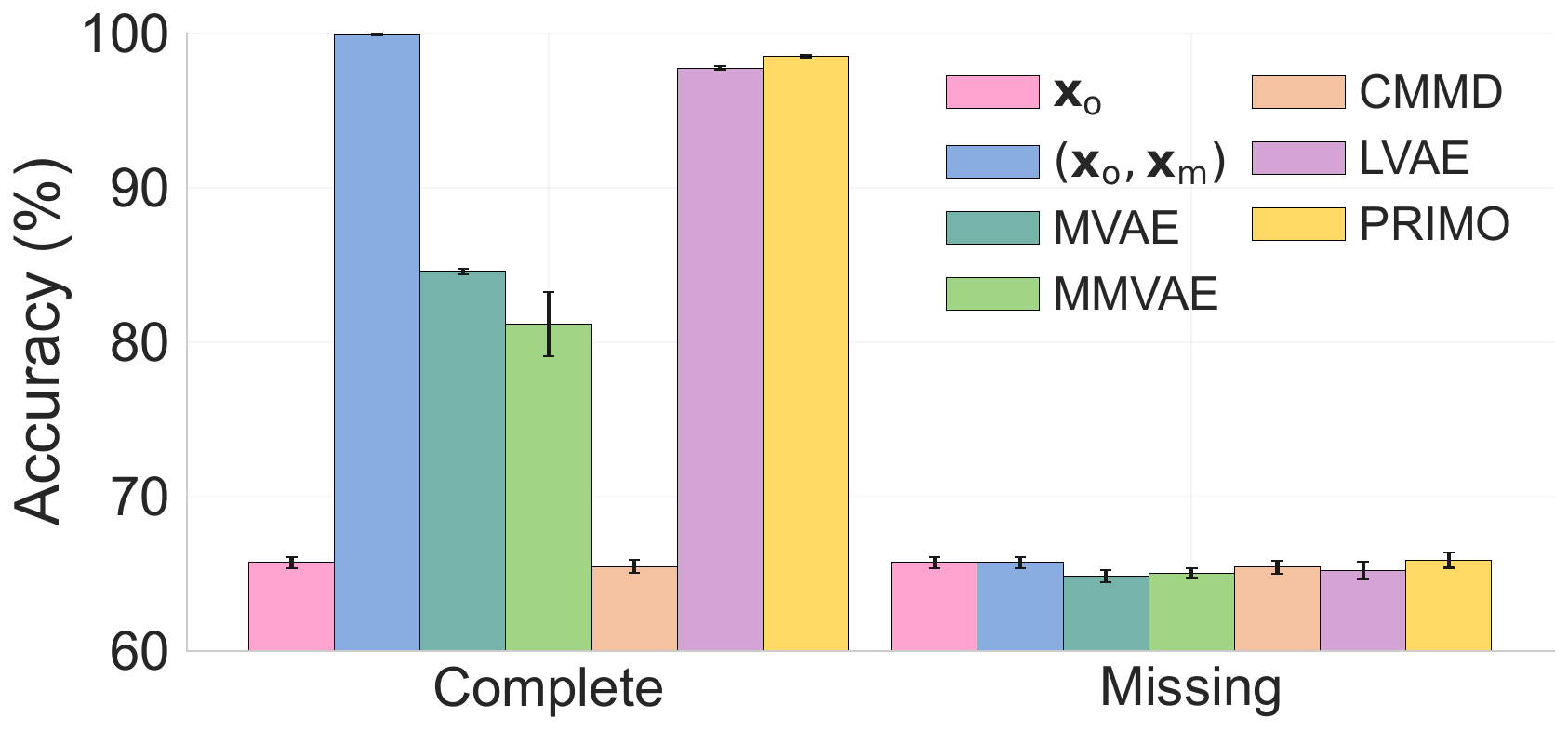}
    % \hfill
    \hspace{0.1in}
    \includegraphics[width=0.4\linewidth]{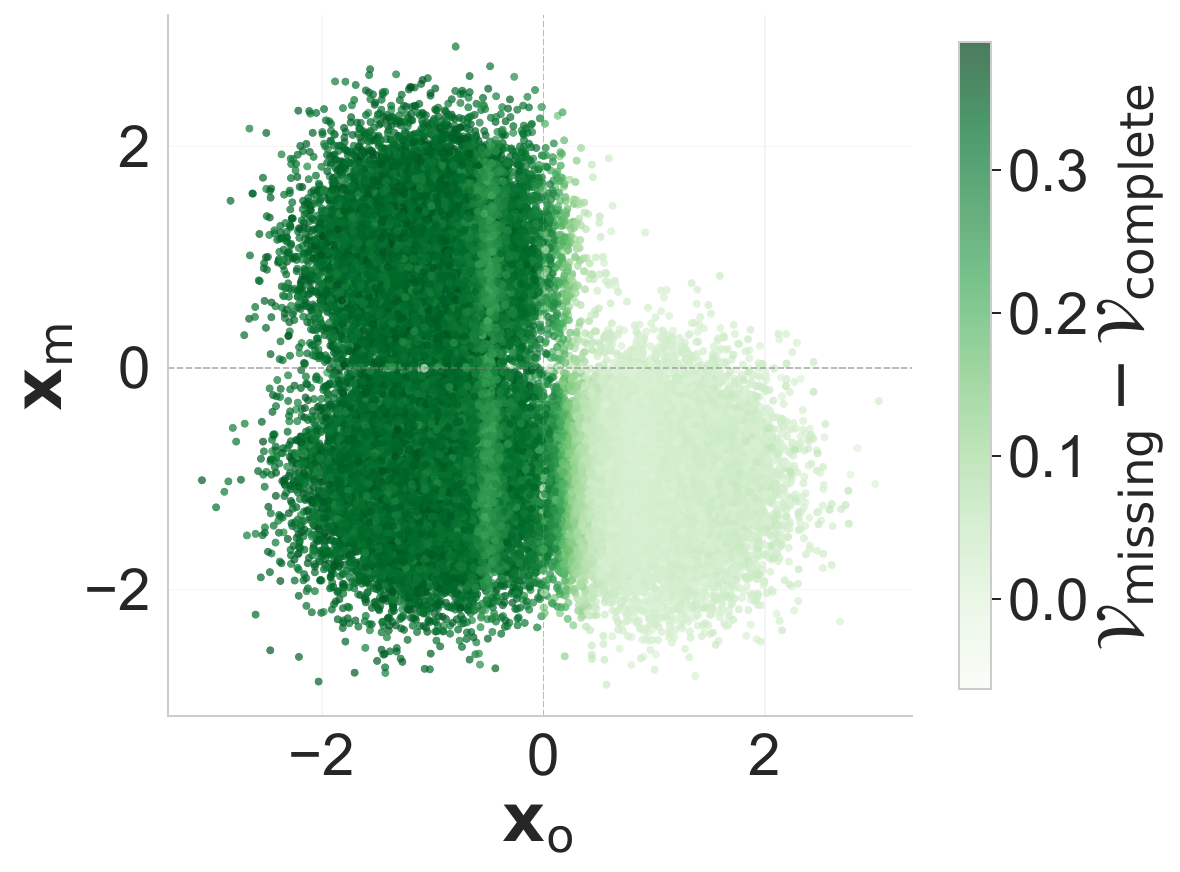}
    \vspace{-0.05in} 
\caption{{\bf Evaluation on the XOR dataset.} {\bf (Left.)} Accuracy under complete and missing-modality inputs. PRIMO matches the unimodal baseline $(\observedmodality)$ when $\missingmodality$ is missing and matches the multimodal baseline $(\observedmodality,\missingmodality)$ when both modalities are observed, outperforming the remaining baselines. {\bf (Right.)} Scatter plot of the predictive impact gap $\mathcal{V}_{\text{missing}}-\mathcal{V}_{\text{complete}}$. The gap is small for examples with $\observedmodality>0$, where the label can be determined by $\observedmodality$ only, and larger for $\observedmodality<0$, where $\missingmodality$ affects the label.}
    \label{fig:xor_results}
\end{figure*}

%% file: figures/figure_2.tex
\begin{figure*}[t]
\centering
\vspace{-0.05in}

\begin{minipage}[t]{0.34\textwidth}
\vspace{0pt}
\centering
\captionof{table}{{\bf Accuracy on AV-MNIST.} We consider audio-missing and vision-missing settings. PRIMO performs comparably to the unimodal baseline that uses the available modality and to the multimodal baseline in both scenarios.\label{tab:avmnist_acc_results}}
\vspace{-0.1in}
\resizebox{\linewidth}{!}{%
\begin{tabular}{lcc}
\toprule
 & Audio & Vision \\
\midrule
$\observedmodality$ & $64.23_{0.17}$ & $40.36_{0.80}$\\
PRIMO & $63.06_{0.72}$ & $37.58_{1.29}$\\
\midrule
$(\observedmodality, \missingmodality)$ & $71.14_{0.42}$ & $71.32_{0.30}$\\
PRIMO  & $68.17_{1.42}$ & $68.27_{1.35}$\\
\bottomrule
\end{tabular}}
\end{minipage}\hfill
\begin{minipage}[t]{0.30\textwidth}
\vspace{0pt}
\centering
\includegraphics[width=\linewidth]{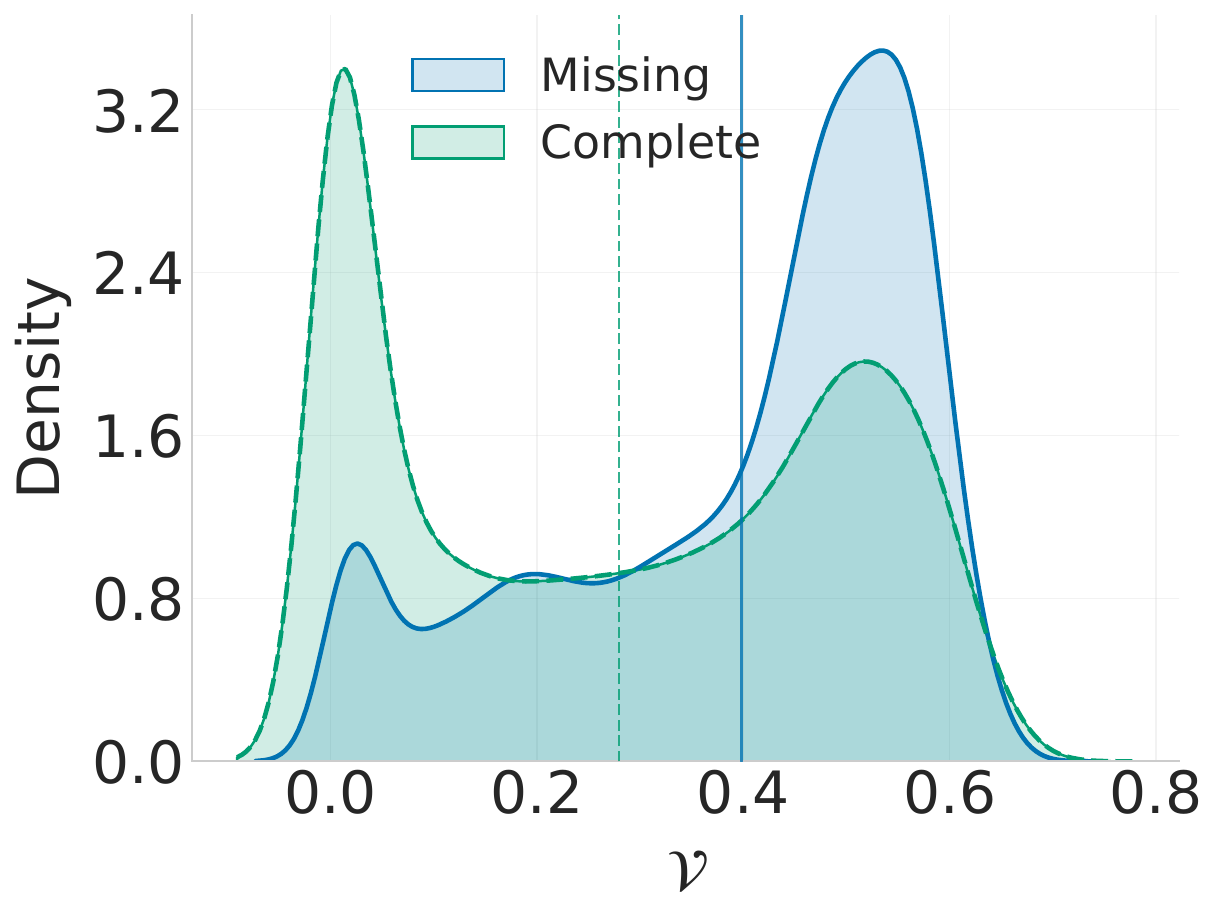}
\captionof{figure}{Distribution of $\mathcal{V}$ when audio is missing. Strong overlap between $\mathcal{V}_{\text{missing}}$ and $\mathcal{V}_{\text{complete}}$ indicates that predictions are often insensitive to the audio modality for those examples.\label{fig:audio_missing_overlap}}

% $\mu_{\text{miss}}~=~0.37_{0.02}$. 
\end{minipage}\hfill
\begin{minipage}[t]{0.30\textwidth}
\vspace{0pt}
\centering
\includegraphics[width=\linewidth]{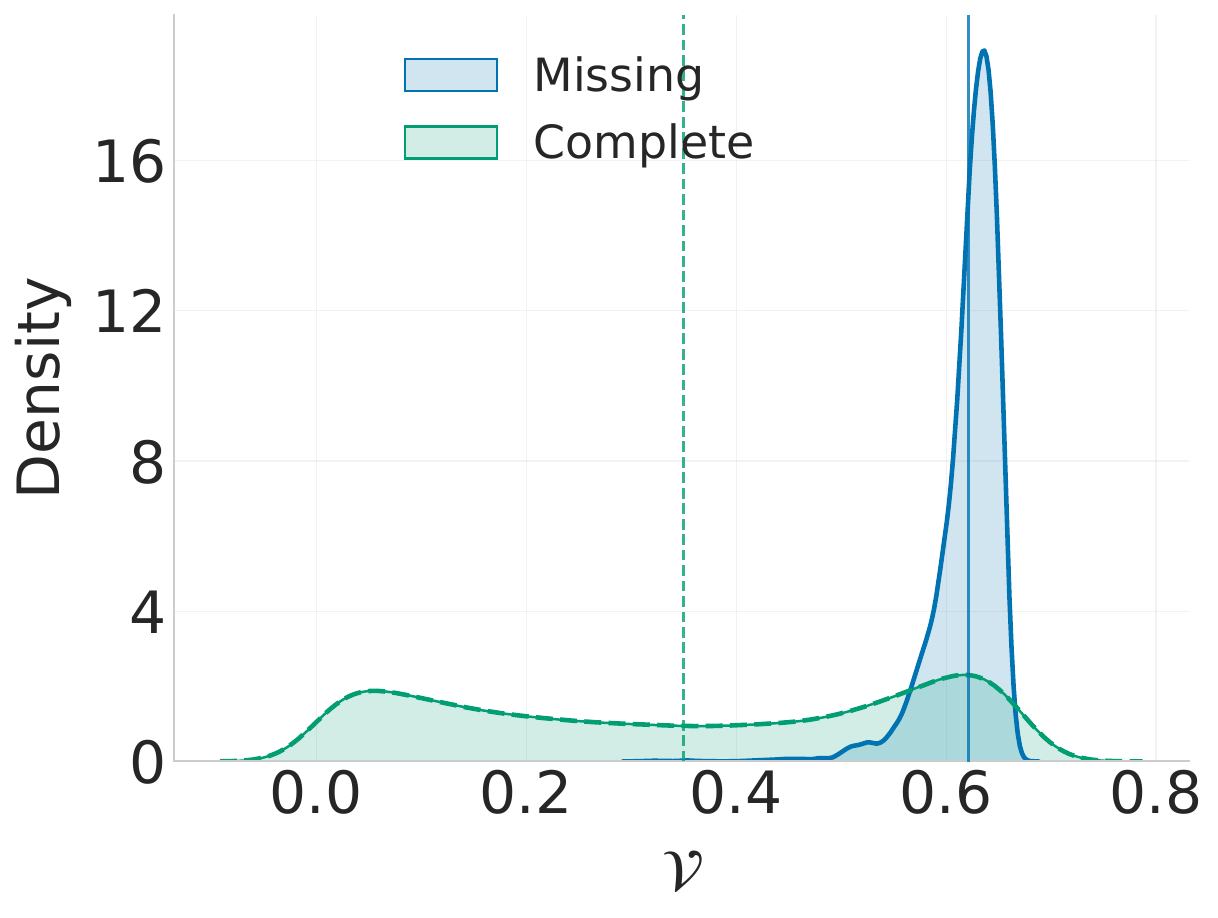}
\vspace{-0.25in}
\captionof{figure}{Distribution of $\mathcal{V}$ when vision is missing. $\mathcal{V}_{\text{missing}}$ is shifted to the right relative to $\mathcal{V}_{\text{complete}}$, indicating greater sensitivity to plausible vision completions.\label{fig:image_missing_overlap}}

% $\mu_{\text{miss}}~=~0.57_{0.02}$.\
\end{minipage}

% \vspace{0.05in}

\pairpanel{0.49\linewidth}{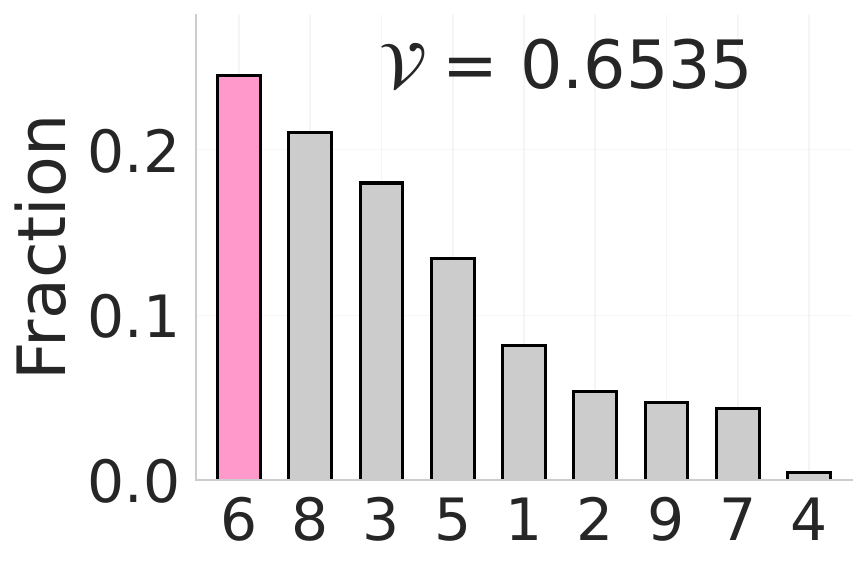}{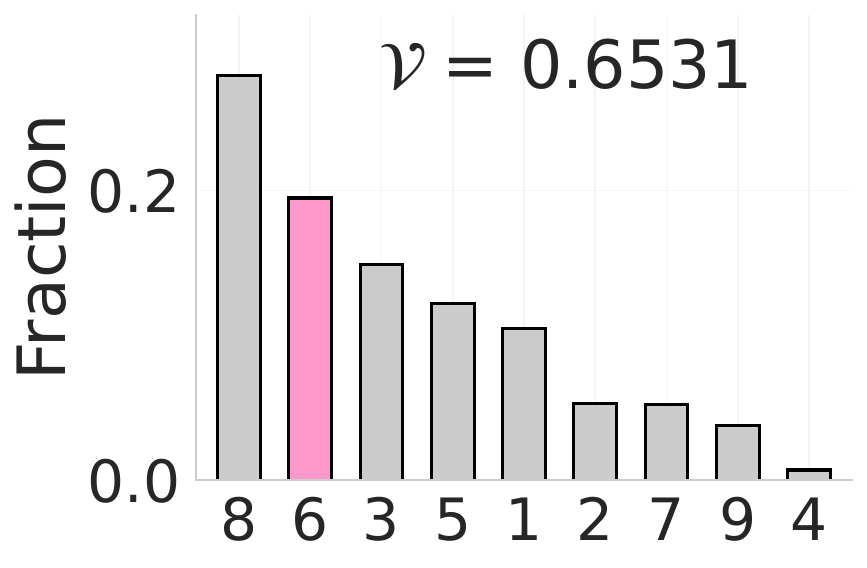}\hfill
\pairpanel{0.49\linewidth}{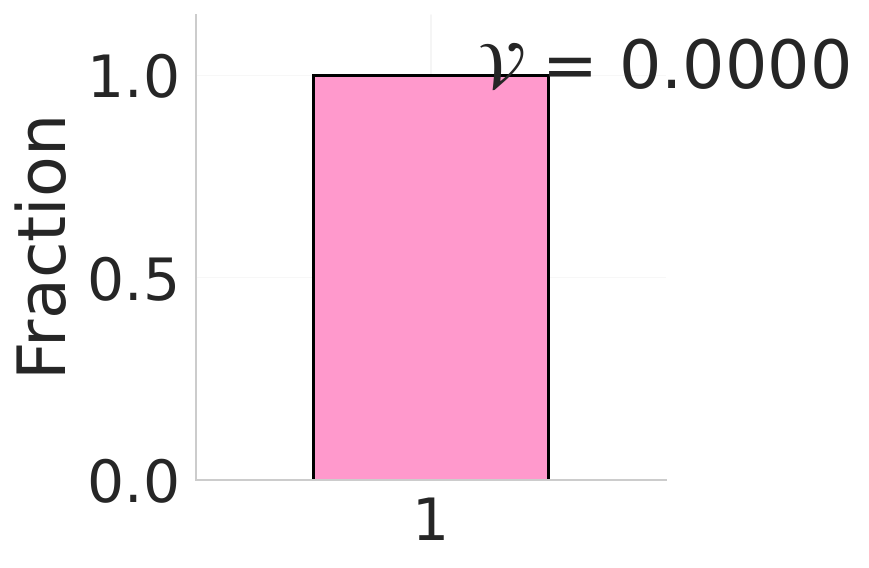}{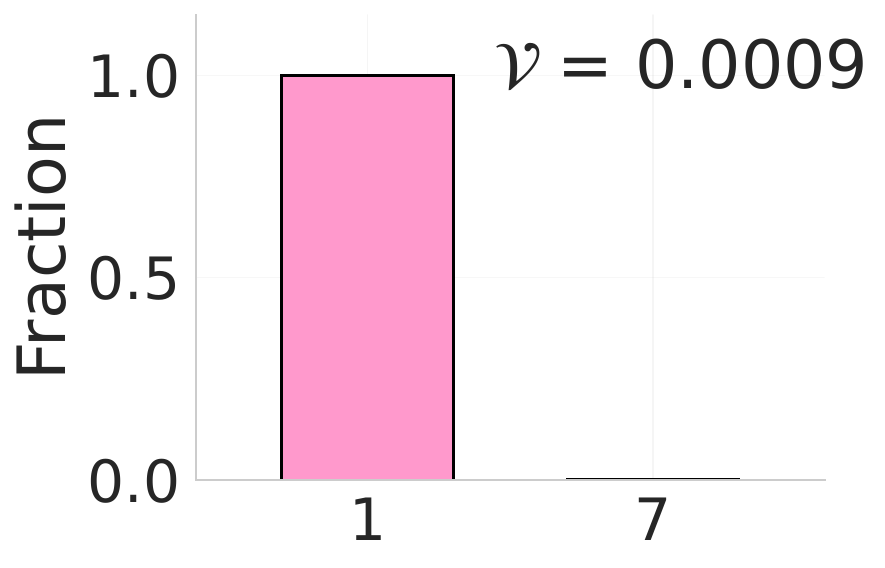}

% \vspace{0.8em}

\pairpanel{0.49\linewidth}{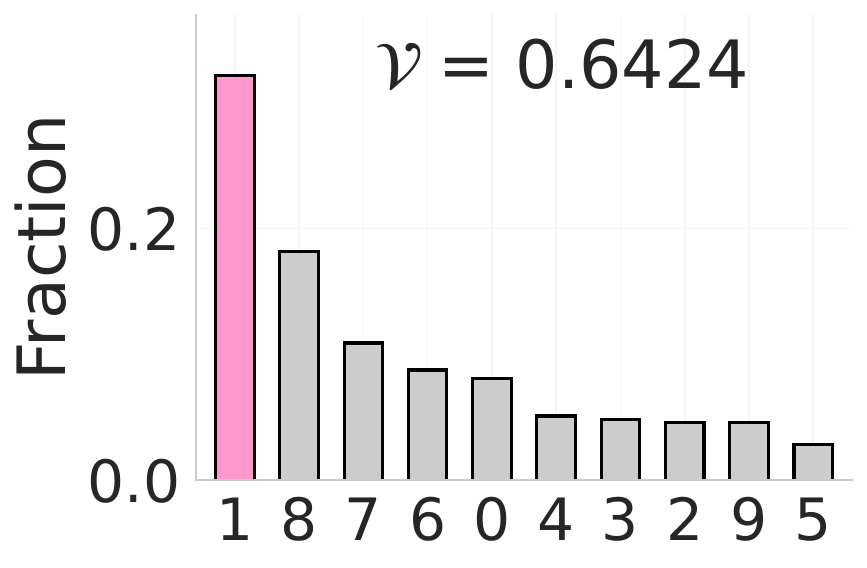}{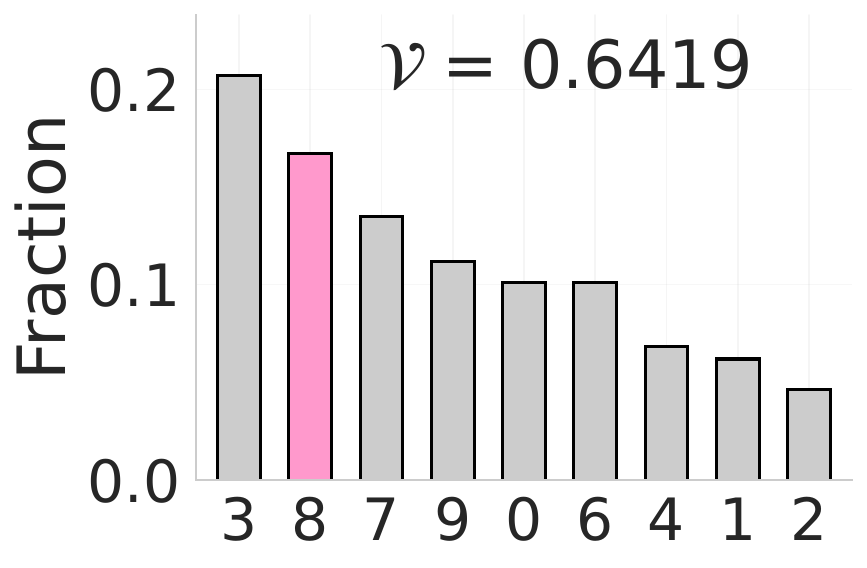}\hfill
\pairpanel{0.49\linewidth}{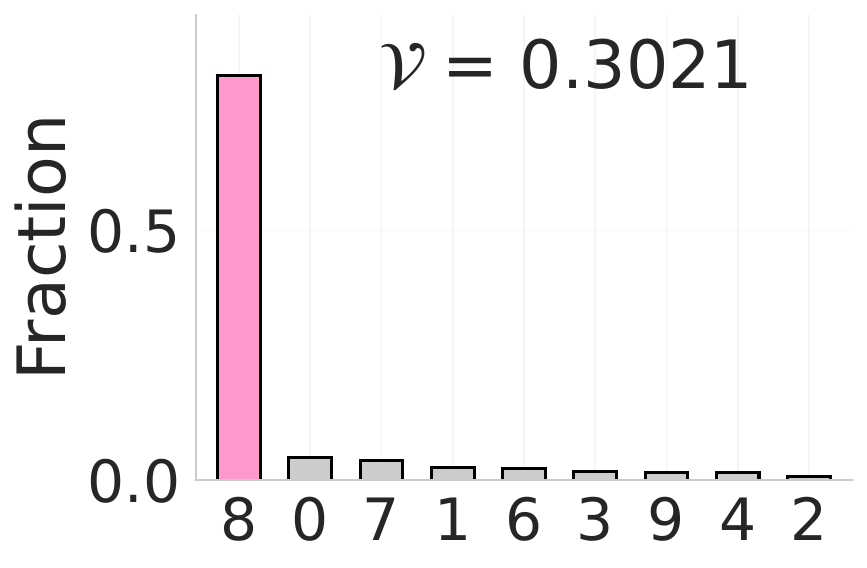}{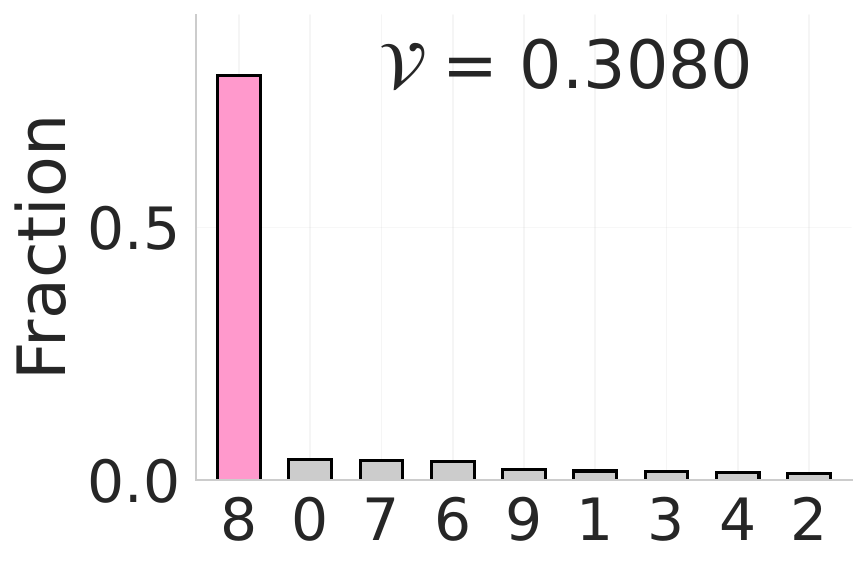}

% \vspace{-0.05in}
\caption{{\bf Qualitative analysis of modality impact on AV-MNIST under audio-missing (top) and vision-missing (bottom)}. We visualize plausible label outcomes induced by varying the latent completion $\latent$. High-$\mathcal{V}$ examples yield multiple plausible label clusters under missingness, while low-$\mathcal{V}$ examples concentrate on a single dominant label.\label{fig:avmnistvisual}}
% \vspace{-0.1in}
\end{figure*}

%% file: figures/figure_3.tex
\begin{figure*}[t]
\centering
\vspace{-0.05in}

\begin{minipage}[t]{0.33\textwidth}
\vspace{0pt}
\centering
\captionof{table}{{\bf MIMIC-III accuracy.} We consider mortality and ICD-9 group prediction under missing and complete modality settings. We report mean and standard deviation across five runs for the unimodal baseline ($\observedmodality$), the multimodal baseline $(\observedmodality,\missingmodality)$, and PRIMO in each setting.\label{tab:mimic_results}}

\vspace{-0.1in}
\resizebox{\linewidth}{!}{%
\begin{tabular}{lccc}
\toprule
 & Mortality & \multicolumn{2}{c}{ICD-9}  \\
\midrule
 & & 140-239 & 460-519 \\
\midrule
$\observedmodality$ & $76.36_{0.01}$ & $91.42_{0.00}$ & $56.22_{0.46}$  \\
PRIMO               & $76.17_{0.07}$ & $91.41_{0.01}$ & $54.95_{1.44}$  \\
\midrule
$(\observedmodality, \missingmodality)$ & $77.89_{0.17}$ & $91.37_{0.10}$ & $68.22_{0.52}$  \\
PRIMO & $77.08_{0.25}$ & $91.41_{0.03}$ & $65.78_{1.08}$  \\
\bottomrule
\end{tabular}}
\end{minipage}
\hfill
% Right: two figures + one caption directly under them
\begin{minipage}[t]{0.66\textwidth}
\vspace{0pt}
\centering

\begin{minipage}[t]{0.51\linewidth}
\vspace{0pt}
\centering
\includegraphics[width=\linewidth]{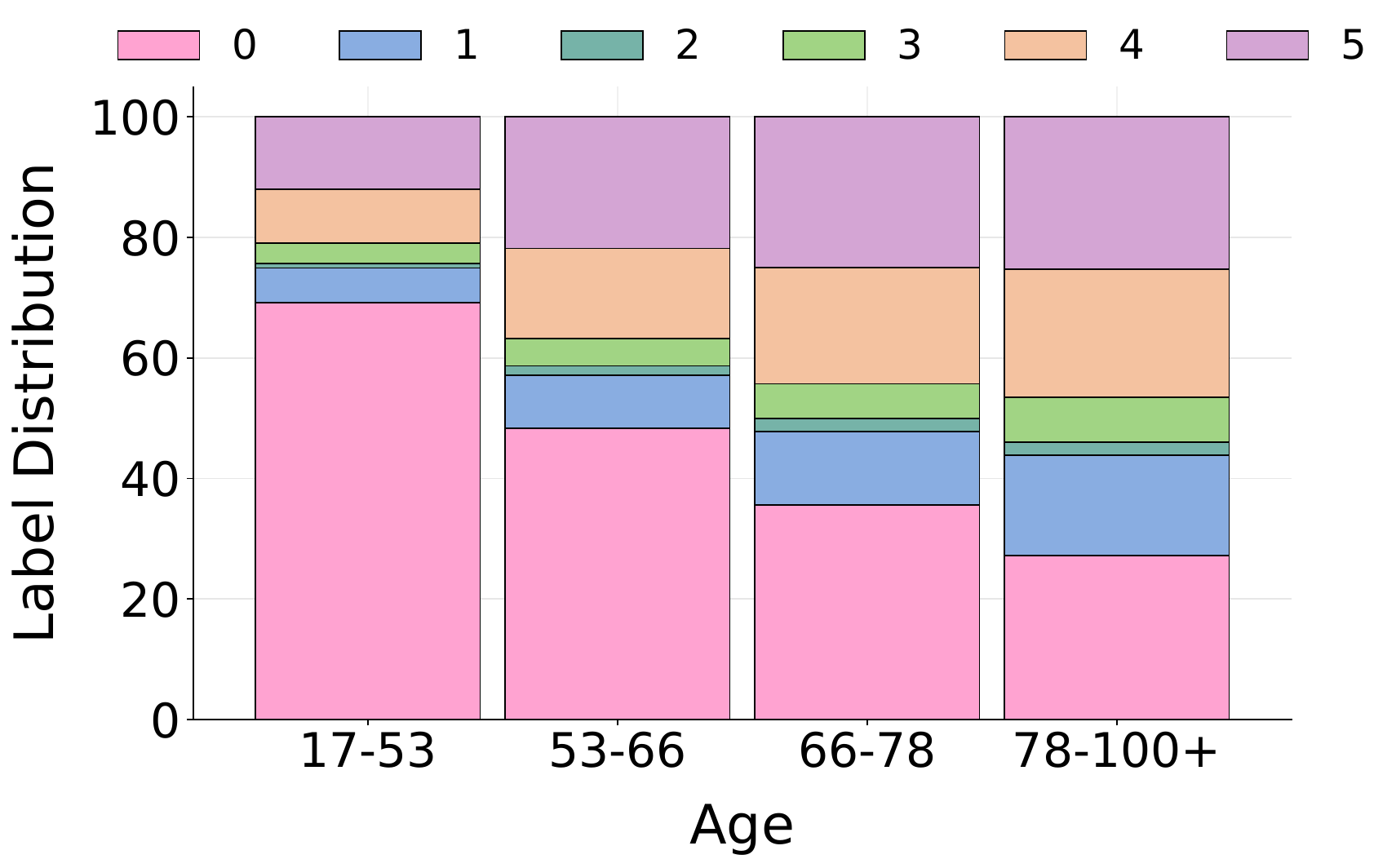}
\end{minipage}\hfill
\begin{minipage}[t]{0.48\linewidth}
\vspace{0pt}
\centering
\includegraphics[width=\linewidth]{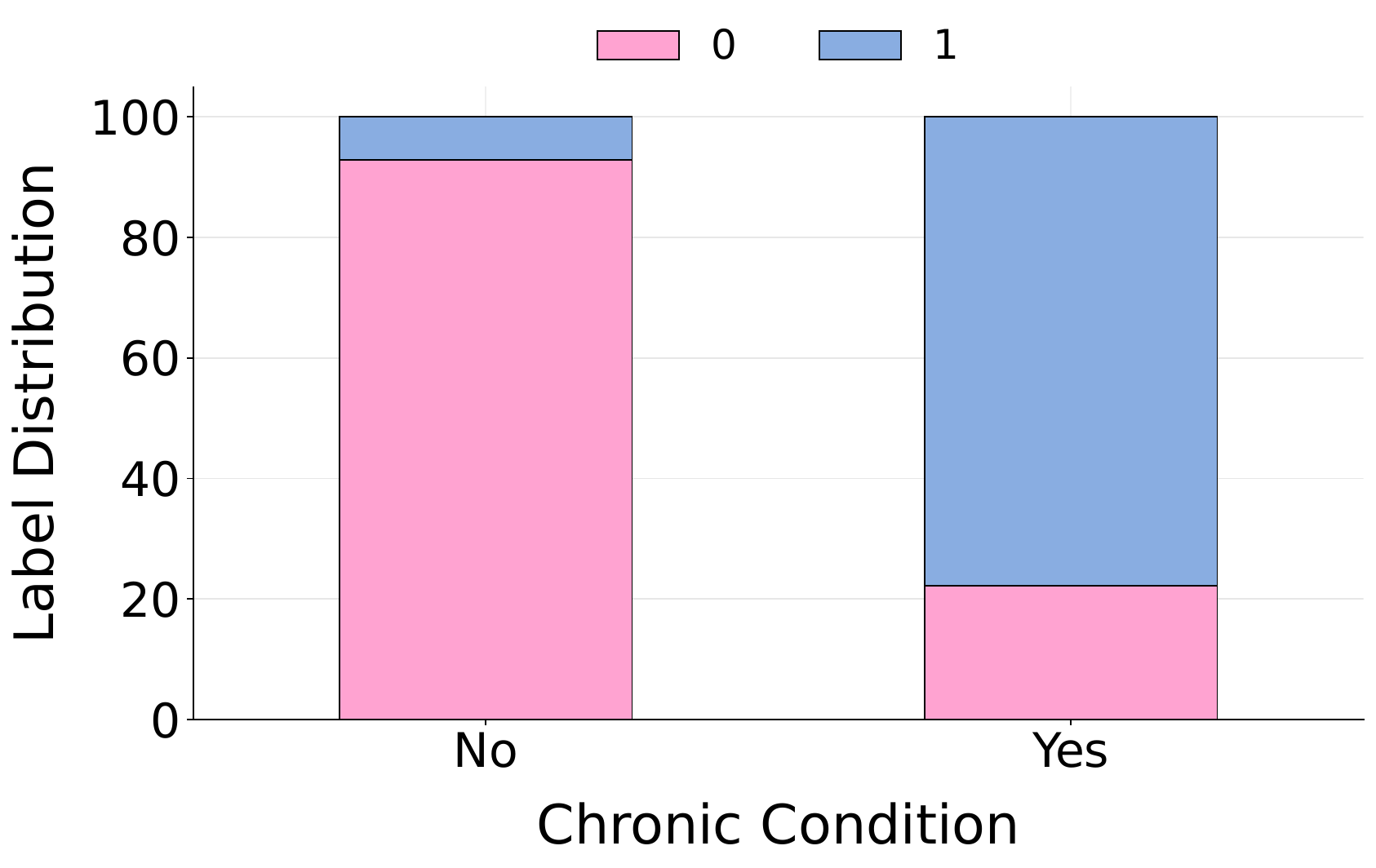}
\end{minipage}
\vspace{-0.05in}
\captionof{figure}{\textbf{(Left)} Cluster-induced plausible label distribution for mortality prediction stratified by age. \textbf{(Right)} Cluster-induced plausible label distribution for ICD-9 neoplasms (140--239) stratified by chronic condition. Distributions are computed from predictions across latent completions of the time-series modality.\label{fig:mortality_var}}

\vspace{0.05in}
\end{minipage}

\hfill
\begin{minipage}[t]{\textwidth}
\vspace{0pt}
\centering
\begin{minipage}[t]{0.32\linewidth}
\vspace{0pt}
\centering
\includegraphics[width=\linewidth]{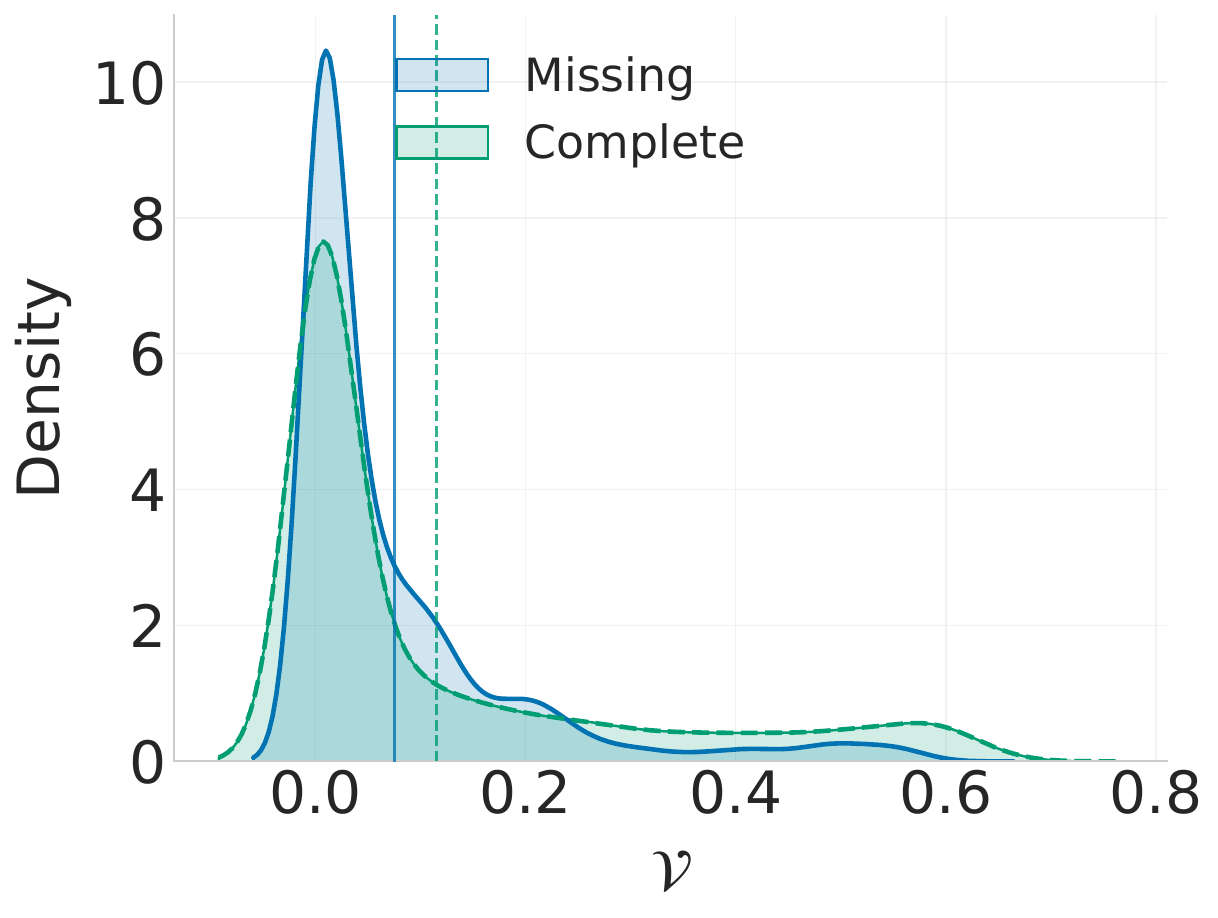}
% \par\small (a) Mortality
\end{minipage}
\begin{minipage}[t]{0.32\linewidth}
\vspace{0pt}
\centering
\includegraphics[width=\linewidth]{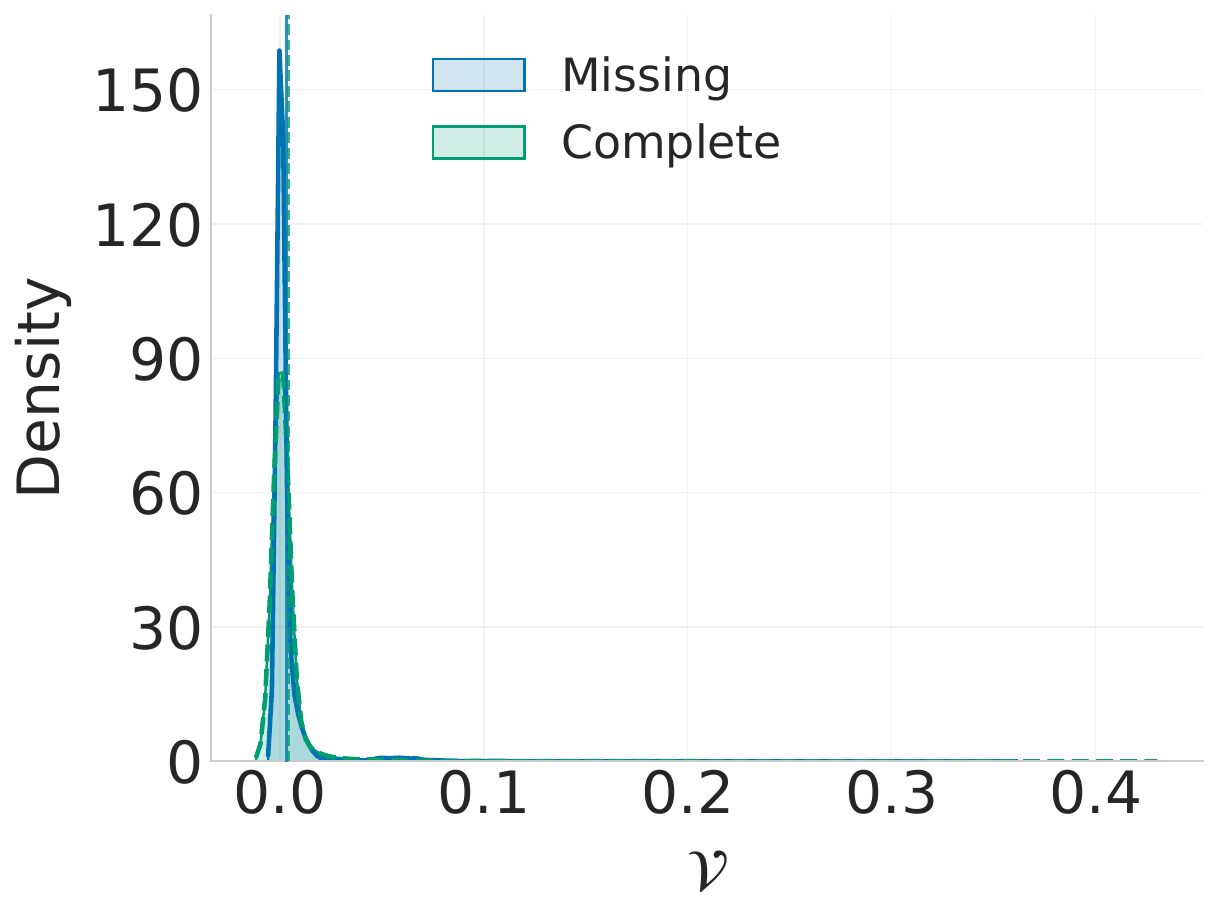}
\end{minipage}
\hfill
\begin{minipage}[t]{0.32\linewidth}
\vspace{0pt}
\centering
\includegraphics[width=\linewidth]{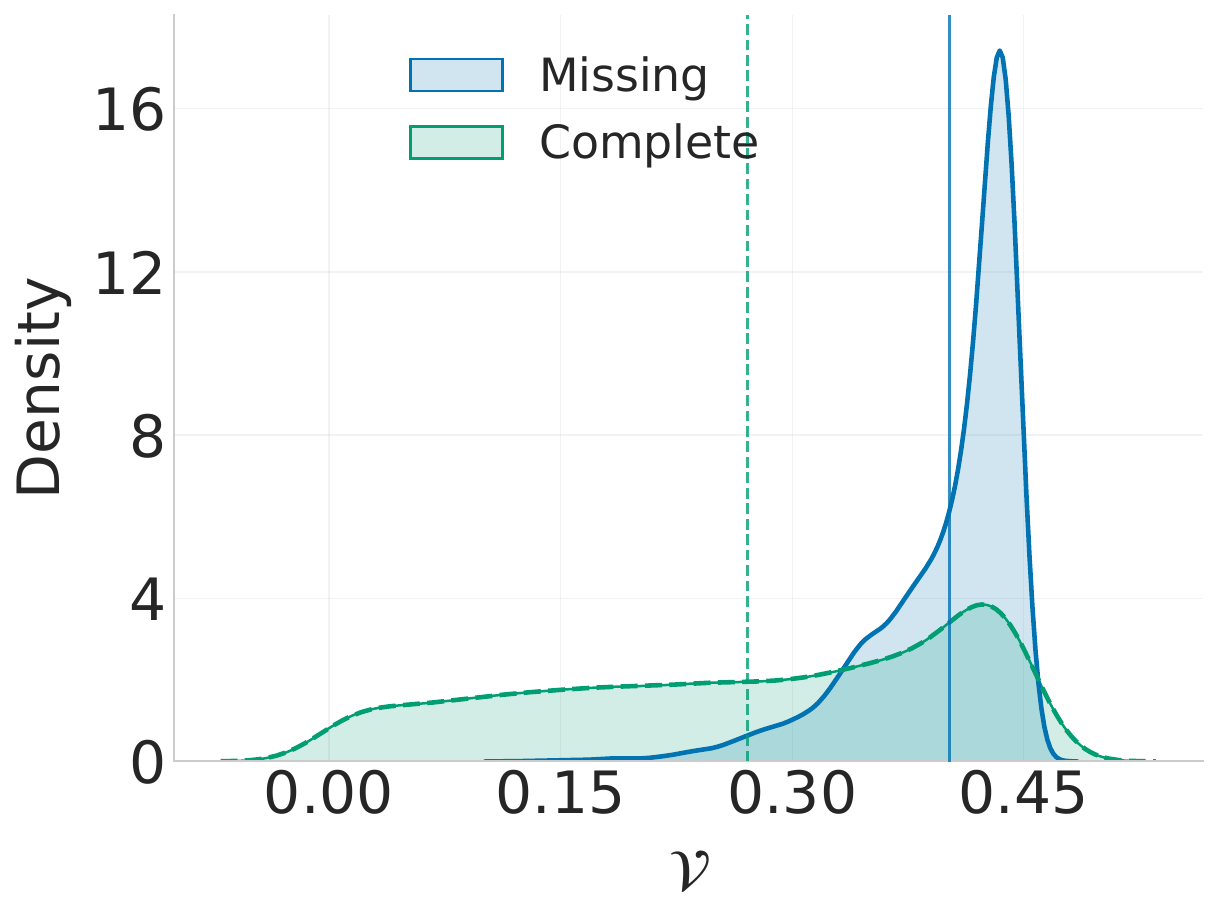}
\end{minipage}

\vspace{-0.1in}
\captionof{figure}{\textbf{Predictive impact under missing and complete time-series modality with sampling $\latent$.}
We compare $\mathcal{V}$ for \textbf{(left)} mortality prediction,  \textbf{(center)} ICD-9 140--239 (neoplasms), and \textbf{(right)} ICD-9 460--519 (respiratory diseases).
The time-series modality has little impact for ICD-9 140--239, but it affects ICD-9 460--519 and mortality prediction.\label{fig:mimic_var}}

\end{minipage}
\vspace{-0.05in}
\end{figure*}

\begin{figure*}
\begin{minipage}[t]{0.32\linewidth}
\vspace{0pt}
\centering
\includegraphics[width=\linewidth]{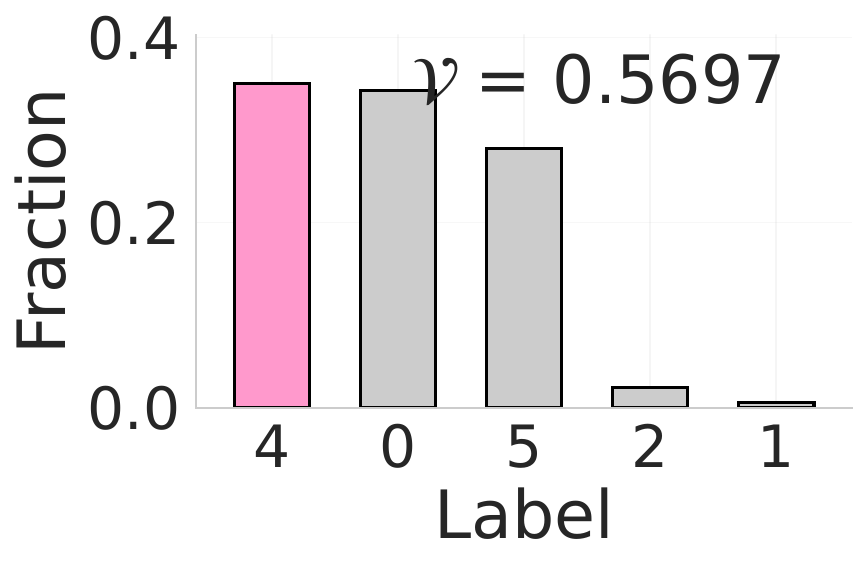}
\par\small (a) Mortality
\end{minipage}\hfill
\begin{minipage}[t]{0.32\linewidth}
\vspace{0pt}
\centering
\includegraphics[width=\linewidth]{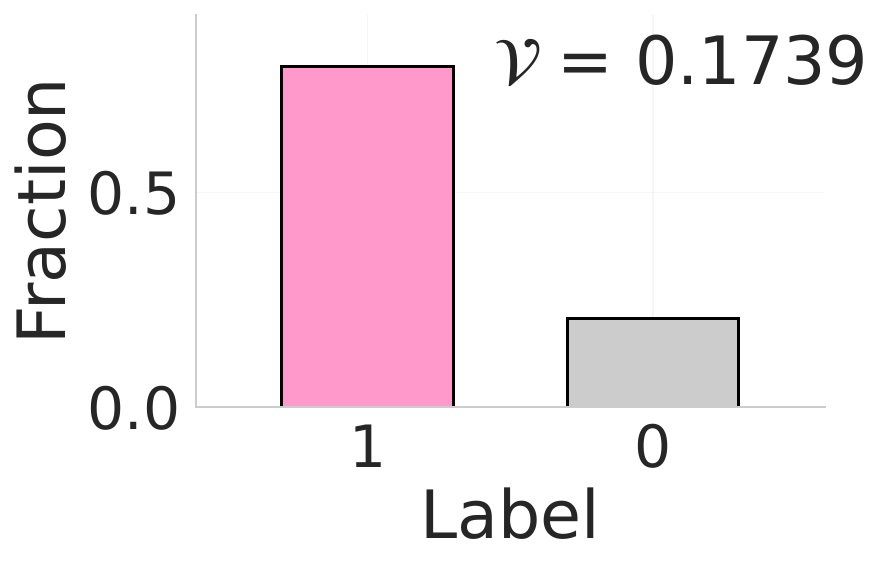}
\par\small (b) ICD-9 (140--239)
\end{minipage}\hfill
\begin{minipage}[t]{0.32\linewidth}
\vspace{0pt}
\centering
\includegraphics[width=\linewidth]{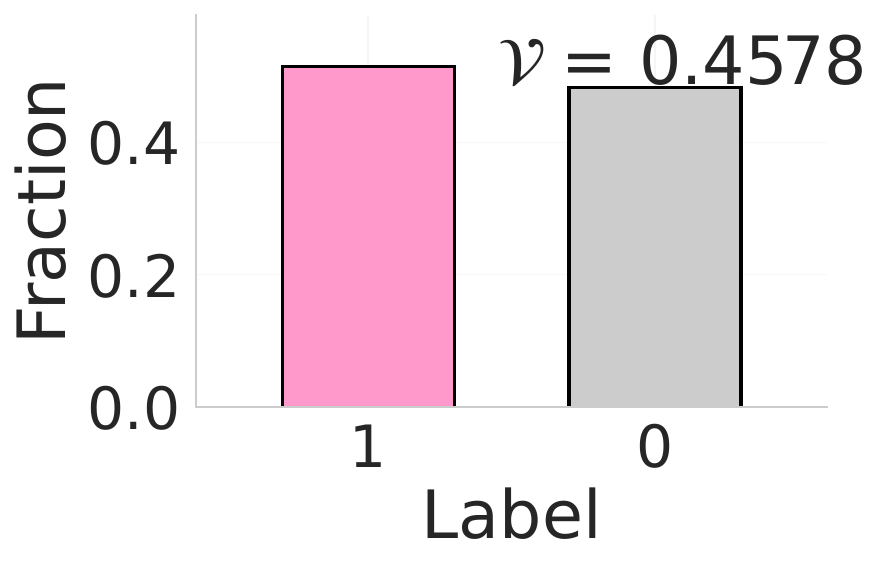}
\par\small (c) ICD-9 (460--519)
\end{minipage}
\vspace{-0.1in}
\captionof{figure}{{\bf Patient-level analysis on MIMIC-III under missing time-series modality.} We report $\mathcal{V}_{\text{missing}}$ as $\mathcal{V}$ and visualize the fraction of clusters assigned to each label. High-$\mathcal{V}$ examples yield clusters spread across multiple labels, indicating ambiguity for (a) mortality risk and (c) respiratory disease diagnosis, whereas (b) neoplasm prediction concentrates on a single label and remains stable.\label{fig:mimic_ind_patients}}

\end{figure*}

%% file: sections/5_conclusion.tex
\section{Limitations and Future Work}
\paragraph{Constraints on validating modality importance.} 
 In practical multimodal settings, we often do not have access to certain modalities at inference time, and in many applications labels may also be missing. This makes it challenging to evaluate whether instance-level modality impact estimates are correct. Our qualitative results on MIMIC-III suggest that modality relevance can vary substantially across tasks and examples within the same dataset. Validating this instance-level modality importance without any ground truth, however, remains an open problem. Incorporating human feedback with automated evaluation protocols would be an interesting direction for future work.

\paragraph{Evaluation with many modalities.} Another practical constraint in multimodal learning is the scarcity of benchmarks with multiple modalities and heterogeneous missingness patterns.
PRIMO can extend to any number of modalities by introducing a latent variable for each potentially missing modality, however, current standard benchmarks focus primarily on audio, vision, and text. This limits evaluation in settings with sensory data, tabular data, and multiple imaging modalities. We hope this work motivates benchmarks with three or more modalities and heterogeneous missingness patterns, enabling more realistic evaluation of imputation-based multimodal learning and instance-level modality importance under incomplete data.

\section{Conclusion}

We propose PRIMO, a supervised latent-variable model for characterizing predictions under plausible completions of a missing modality. PRIMO supports both complete and missing-modality settings, and achieves performance comparable to unimodal baselines when a modality is missing and multimodal baselines when all modalities are observed. Beyond predictive performance, PRIMO provides instance-level estimates of how missing modalities affect predictions across datasets and modality combinations. We find that modality contributions vary across tasks and across examples within the same dataset. These results highlight the heterogeneity of multimodal datasets. PRIMO provides a principled way to capture this heterogeneity as modality availability and relevance change.

\section*{Acknowledgement}
This work was supported by the Institute of Information \& Communications Technology Planning \& Evaluation (IITP) with a grant funded by the Ministry of Science and ICT (MSIT) of the Republic of Korea in connection with the Global AI Frontier Lab International Collaborative Research, Samsung Advanced Institute of Technology (under the project Next Generation Deep Learning: From Pattern Recognition to AI), National Science Foundation (NSF) award No. 1922658, Center for Advanced Imaging Innovation and Research (CAI2R), National Center for Biomedical Imaging and Bioengineering operated by NYU Langone Health, and National Institute of Biomedical Imaging and Bioengineering through award number P41EB017183. The computational requirements for this work were supported by NYU IT High Performance Computing resources, services, and staff expertise and NYU
Langone High Performance Computing Core’s resources and personnel. This content is solely the
responsibility of the authors and does not represent the views of the funding agencies.

%% file: sections/6_appendix.tex
\clearpage
\onecolumn
\appendix

{\bf Organization.} The Appendix includes ELBO derivations (\Cref{sec:elboderivation}), detailed related work on multimodal learning with missing modalities (\Cref{appendix:relatedwork}), and the experimental setup with additional results (\Cref{appendixsection:experiments}).

\section{ELBO derivations}\label{sec:elboderivation}

\subsection{Complete modalities ($\mathcal{D}_{\text{complete}}$)}
\label{sec:elboderivation_complete}

When both modalities $\observedmodality$ and $\missingmodality$ are available, we maximize the conditional log-likelihood $\log p(\labely \mid \observedmodality, \missingmodality)$. Following the dependencies from our graphical model in \Cref{fig:dgp}, the joint distribution of the label and latent variable factorizes as $p(\labely, \latent \mid \observedmodality, \missingmodality)~= p_\theta(\labely \mid \observedmodality, \latent) p_\omega(\latent \mid \observedmodality, \missingmodality)$, where $\labely \perp \missingmodality \mid \{\latent, \observedmodality\}$. By introducing the variational posterior $q_\phi(\latent \mid \observedmodality, \missingmodality, \labely)$, we derive the ELBO as follows:
\begin{equation}
\begin{aligned}
    \log p(\labely \mid \observedmodality, \missingmodality) 
    &= \log \int p_\theta(\labely \mid \observedmodality, \latent) p_\omega(\latent \mid \observedmodality, \missingmodality) \, d\latent \nonumber \\
    &= \log \int q_\phi(\latent \mid \observedmodality, \missingmodality, \labely) \frac{p_\theta(\labely \mid \observedmodality, \latent) p_\omega(\latent \mid \observedmodality, \missingmodality)}{q_\phi(\latent \mid \observedmodality, \missingmodality, \labely)} \, d\latent \nonumber \\
    &\geq \int q_\phi(\latent \mid \observedmodality, \missingmodality, \labely) \log \left( \frac{p_\theta(\labely \mid \observedmodality, \latent) p_\omega(\latent \mid \observedmodality, \missingmodality)}{q_\phi(\latent \mid \observedmodality, \missingmodality, \labely)} \right) \, d\latent \quad \text{(Jensen's Inequality)} \nonumber \\
    &= \mathbb{E}_{q_\phi} \left[ \log p_\theta(\labely \mid \observedmodality, \latent) \right] + \mathbb{E}_{q_\phi} \left[ \log \frac{p_\omega(\latent \mid \observedmodality, \missingmodality)}{q_\phi(\latent \mid \observedmodality, \missingmodality, \labely)} \right] \nonumber \\
    &= \mathbb{E}_{q_\phi(\latent \mid \observedmodality, \missingmodality, \labely)} \left[ \log p_\theta(\labely \mid \observedmodality, \latent) \right] - \text{KL}\left( q_\phi(\latent \mid \observedmodality, \missingmodality, \labely) \,\Vert\, p_\omega(\latent \mid \observedmodality, \missingmodality) \right).
\end{aligned}
\end{equation}

\subsection{Missing modality ($\mathcal{D}_{\text{missing}}$)}
\label{sec:elboderivation_missing}

In the case where $\missingmodality$ is unavailable, we maximize $\log p(\labely \mid \observedmodality)$. The dashed edge in \Cref{fig:dgp} indicates a correlation between $\observedmodality$ and $\missingmodality$, implying that $\observedmodality$ carries information about the missing modality. We thus utilize a conditional prior $p_\omega(\latent \mid \observedmodality)$ to infer $\latent$. Using the variational posterior $q_\phi(\latent \mid \observedmodality, \labely)$, we derive the ELBO as:
\begin{equation}
\begin{aligned}
    \log p(\labely \mid \observedmodality) 
    &= \log \int p_\theta(\labely \mid \observedmodality, \latent) p_\omega(\latent \mid \observedmodality) \, d\latent \nonumber \\
    &= \log \int q_\phi(\latent \mid \observedmodality, \labely) \frac{p_\theta(\labely \mid \observedmodality, \latent) p_\omega(\latent \mid \observedmodality)}{q_\phi(\latent \mid \observedmodality, \labely)} \, d\latent \nonumber \\
    &\geq \int q_\phi(\latent \mid \observedmodality, \labely) \log \left( \frac{p_\theta(\labely \mid \observedmodality, \latent) p_\omega(\latent \mid \observedmodality)}{q_\phi(\latent \mid \observedmodality, \labely)} \right) \, d\latent \quad \text{(Jensen's Inequality)} \nonumber \\
    &= \mathbb{E}_{q_\phi} \left[ \log p_\theta(\labely \mid \observedmodality, \latent) \right] - \mathbb{E}_{q_\phi} \left[ \log \frac{q_\phi(\latent \mid \observedmodality, \labely)}{p_\omega(\latent \mid \observedmodality)} \right] \nonumber \\
    &= \mathbb{E}_{q_\phi(\latent \mid \observedmodality, \labely)} \left[ \log p_\theta(\labely \mid \observedmodality, \latent) \right] - \text{KL}\left( q_\phi(\latent \mid \observedmodality, \labely) \,\Vert\, p_\omega(\latent \mid \observedmodality) \right).
\end{aligned}
\end{equation}

\section{Related Work on Multimodal learning with Missing Modalities}\label{appendix:relatedwork}

Many prior multimodal VAE studies~\citep{suzuki_joint_2017,vedantam_generative_2018, tsai_learning_2019, shi_variational_2019,sutter_generalized_2021,gong_variational_2021,joy_learning_2022, palumbo_mmvae_2023}
 focus on generative modeling by optimizing the marginal likelihood $\max_\theta \mathbb{E}_{q(\latent \mid \observedmodality)} \left[ \log p_\theta(\observedmodality \mid \latent) \right] - \beta \text{KL}(q(\latent\mid \observedmodality) \| p(\latent))$
rather than improving discriminative performance under missing modalities.
JMVAE~\citep{suzuki_joint_2017} and tELBO~\citep{vedantam_generative_2018} 
model the joint distribution $p(\firstmodality, \secondmodality)$. 
These methods use paired multimodal examples during training to learn 
an inference network $p_\theta(\latent \mid \firstmodality, \secondmodality)$ conditioned on all modalities. 
To scale beyond two modalities, MVAE~\citep{wu_multimodal_2018} combines modality-specific posteriors using product-of-experts $q_\phi(\latent\mid \mathbf{x}_{1:M}) \propto \prod_m q_{\phi_m}(\latent\mid \mathbf{x}_m)$,
while MMVAE~\citep{shi_variational_2019} uses a mixture of experts $q_\phi(\latent\mid \mathbf{x}_{1:M})~=~\sum_m \pi_m\, q_{\phi_m}(\latent\mid \mathbf{x}_m)$. 
MoPoE~\citep{sutter_generalized_2021} further generalizes these objectives using a mixture of product of experts. These methods commonly rely on sub-sampling modality subsets, 
which imposes an undesirable upper bound on the multimodal ELBO~\citep{daunhawer_limitations_2022}.
Similarly, while mmJSD~\citep{sutter_multimodal_2020} and MMVAE+~\citep{palumbo_mmvae_2023}
seperated modality-specific and shared latent spaces, they fundamentally optimize
$\log p(\observedmodality)$ and use fully paired multimodal data for training. 

Optimizing a generative ELBO learns a $\latent$ that captures the input variation, 
which might not align with the optimal class decision boundary required for modeling $p(\labely \mid \cdot)$.
CMMD~\citep{mancisidor_discriminative_2024} took a step in this direction 
by incorporating a discrimiantive component into the
multimodal latent framework, 
but assumes fully observed multimodal data during training. 
Using fully paired multimodal data in this setup creates a train-test mismatch,
the posterior learnt during training $q(\latent \mid \observedmodality, \missingmodality, \labely)$
has access to $\missingmodality$ but at test time only $\observedmodality$ is available. MEME~\citep{joy_learning_2022} and VSVAE~\citep{gong_variational_2021} considered
partial multimodal missignenss (only a subset of training examples contains
all modalities), but they were limited to generative modeling. We argue that
both these 
settings are crucial because we need to align the $\latent$ when the modality is 
observed and missing. 
PRIMO explicitly addresses the above limitations and is designed for discriminative setups when modalities are partially observed during both
training and testing.

\section{Additional Experiments}\label{appendixsection:experiments}

This section summarizes the hyperparameters and architectural details (see \Cref{sec:experimental_setup}) with additional experimental results (see \Cref{sec:additional_results}).

\subsection{Experimental Setup}\label{sec:experimental_setup}

\textbf{XOR.} We generate a synthetic 2D XOR classification task with 40,000 samples drawn from four Gaussian clusters centered at $(\pm 1, \pm 1)$ with standard deviation 0.5. Three quadrants are used to create a dataset where $\missingmodality$ provides non-redundant information for classification. We use a 70/30 train/test split and randomly drop the second modality $\missingmodality$ for 50\% of training examples. Each modality is a single scalar feature encoded through a shared two-layer MLP architecture. The prior and posterior are two-layer MLPs with hidden dimension $128$, projecting to a two-dimensional latent space. We train with AdamW (learning rate $1\times10^{-3}$ and weight decay $1\times10^{-4}$). For evaluation, we use $200$ Monte Carlo samples. Results are averaged over four random seeds.

\textbf{AVMNIST.}
We use modality-specific LeNet encoders (three layers for vision and five layers for audio), each followed by a linear projection to a $128$-dimensional latent space. The prior and posterior are implemented as MLPs, with the posterior conditioned on both modalities and the label, and the prior conditioned on the fused representation. We share the same prior/posterior parameters across complete and missing-modality scenarios; when a modality is missing, its representation is zeroed out before fusion. We train with AdamW (learning rate $5\times10^{-4}$). At evaluation, results are estimated using $2000$ Monte Carlo samples.

\textbf{MIMIC-III.}
We use an $80/10/10$ train/validation/test split and randomly drop time-series modality for $50\%$ of training examples. Static features are encoded with a two-layer MLP and time-series features with a GRU; both are projected to a 64-dimensional latent space. The prior and posterior are MLPs conditioned on the available modalities (and the label for the posterior). As in AVMNIST, we share prior/posterior parameters across complete and missing-modality settings, and zero out the representation of any missing modality. We train with AdamW (learning rate $5\times10^{-4}$). For evaluation, we use $500$ Monte Carlo samples.

\subsection{Additional Results}\label{sec:additional_results}
\begin{figure}[t]
    \centering

    \includegraphics[width=\linewidth]{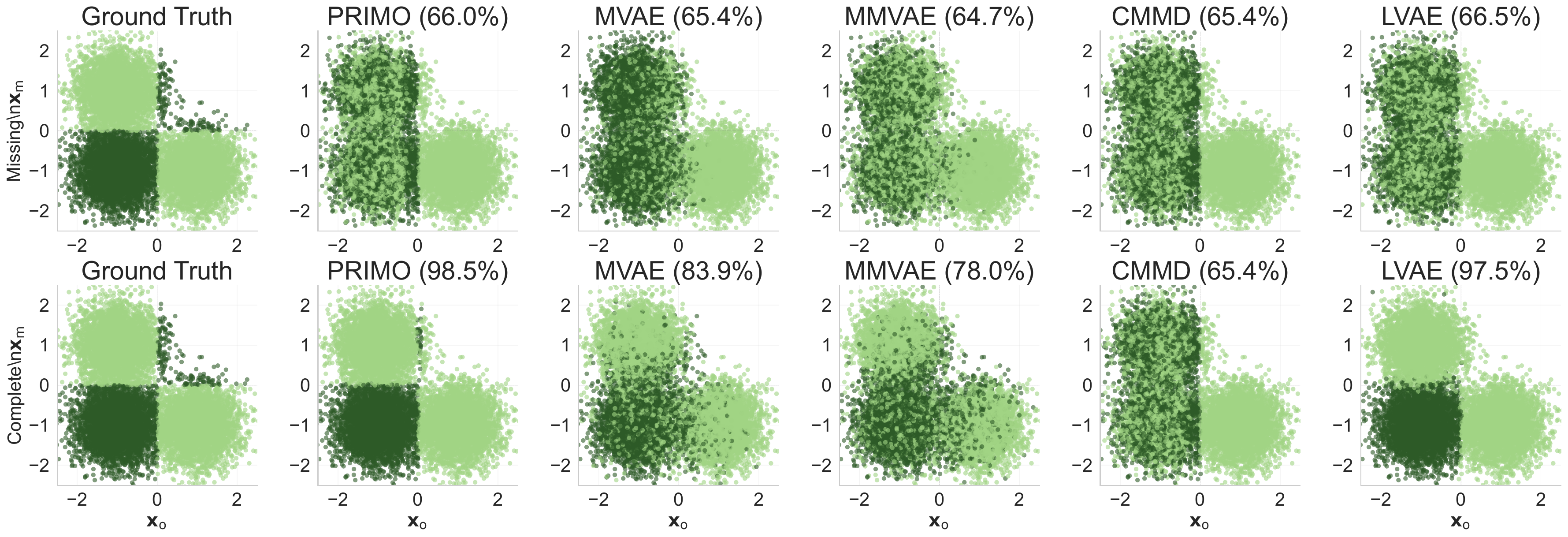}
\caption{{\bf XOR predictions under missing and complete modalities.} Each column shows a method and each row corresponds to the modality-availability setting (top: $\missingmodality$ missing, bottom: complete). For each input $\observedmodality$, we sample latent completions and visualize the induced distribution over predicted labels; points are colored by the predicted class. Methods that capture label-relevant uncertainty produce multiple plausible labels in regions where the XOR label depends on $\missingmodality$ (e.g., $\observedmodality<0$), while predictions concentrate on a single label when $\observedmodality$ is sufficient (e.g., $\observedmodality>0$). Accuracies are shown in the column headers.}
    \label{fig:xor_predictions}

    \vspace{1em}

    \includegraphics[width=\linewidth]{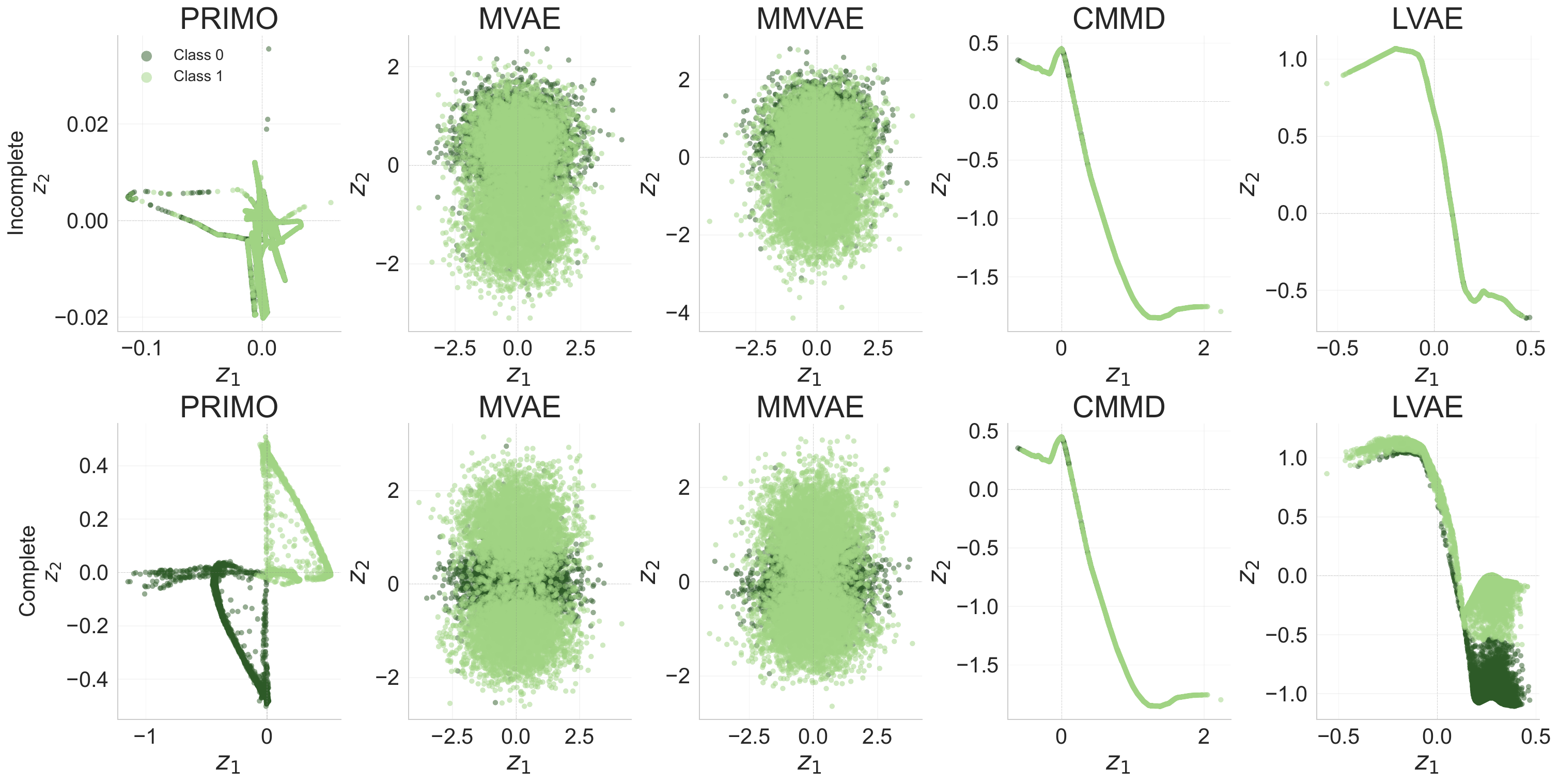}
\caption{{\bf XOR latent-space structure across methods.} We visualize the 2D latent representations used by each method for incomplete inputs (top row) and complete inputs (bottom row). Points are colored by the predicted class under the corresponding latent sample. }

    \label{fig:xor_latent}

\end{figure}

\begin{figure*}[htbp]
\centering
% Top Row: High
\begin{minipage}[t]{0.32\linewidth}
\centering
\includegraphics[width=\linewidth]{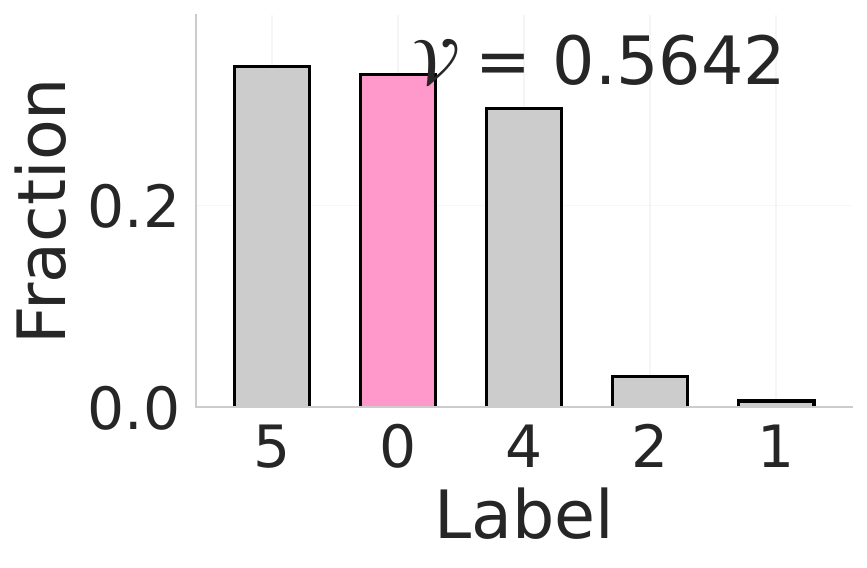}
\end{minipage}\hfill
\begin{minipage}[t]{0.32\linewidth}
\centering
\includegraphics[width=\linewidth]{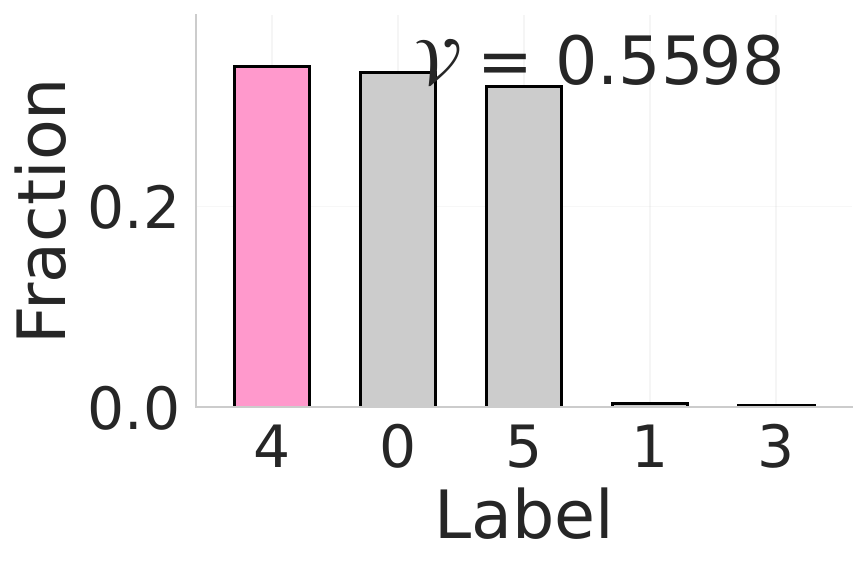}
\end{minipage}\hfill
\begin{minipage}[t]{0.32\linewidth}
\centering
\includegraphics[width=\linewidth]{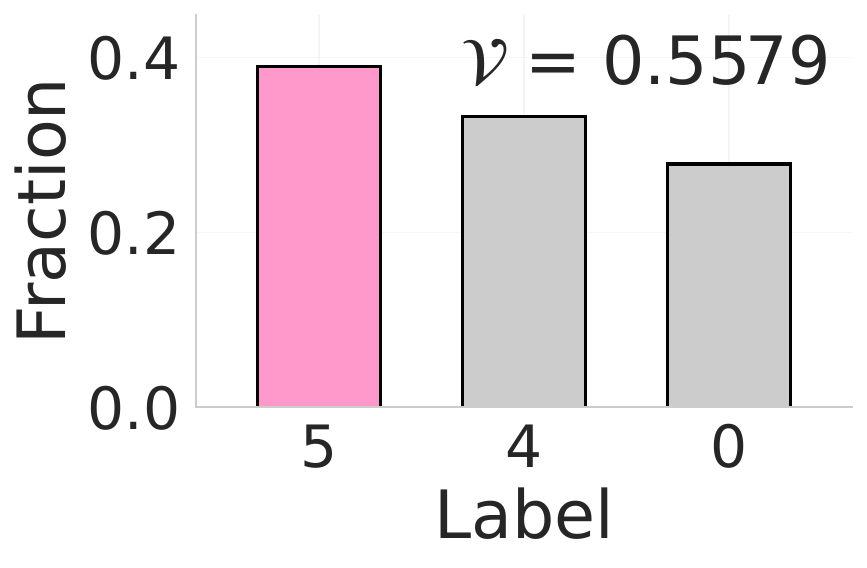}
\end{minipage}

\vspace{1em} % Space between rows

% Bottom Row: Low
\begin{minipage}[t]{0.32\linewidth}
\centering
\includegraphics[width=\linewidth]{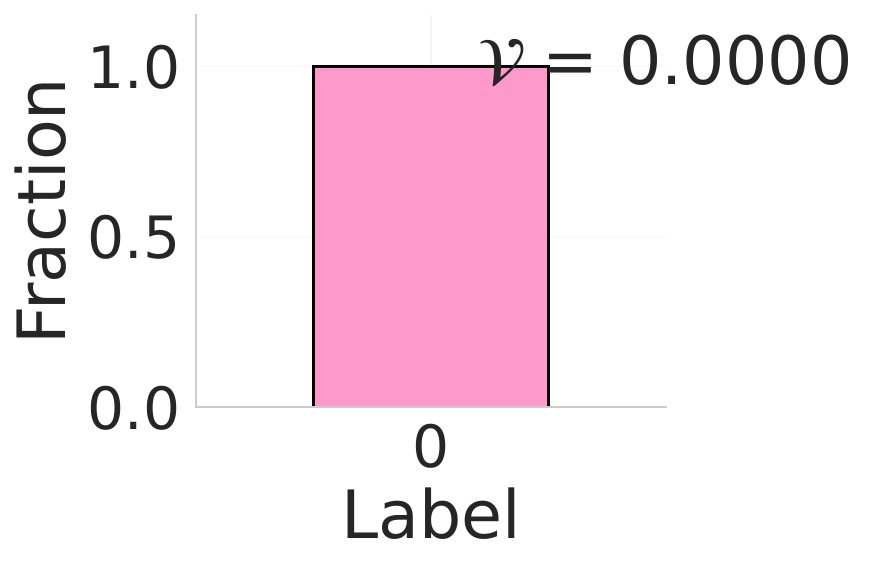}
\end{minipage}\hfill
\begin{minipage}[t]{0.32\linewidth}
\centering
\includegraphics[width=\linewidth]{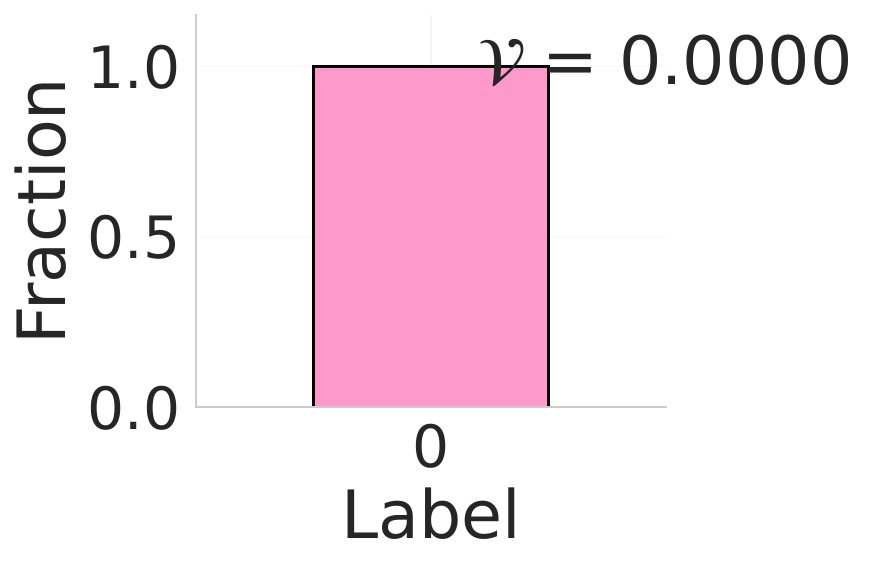}
\end{minipage}\hfill
\begin{minipage}[t]{0.32\linewidth}
\centering
\includegraphics[width=\linewidth]{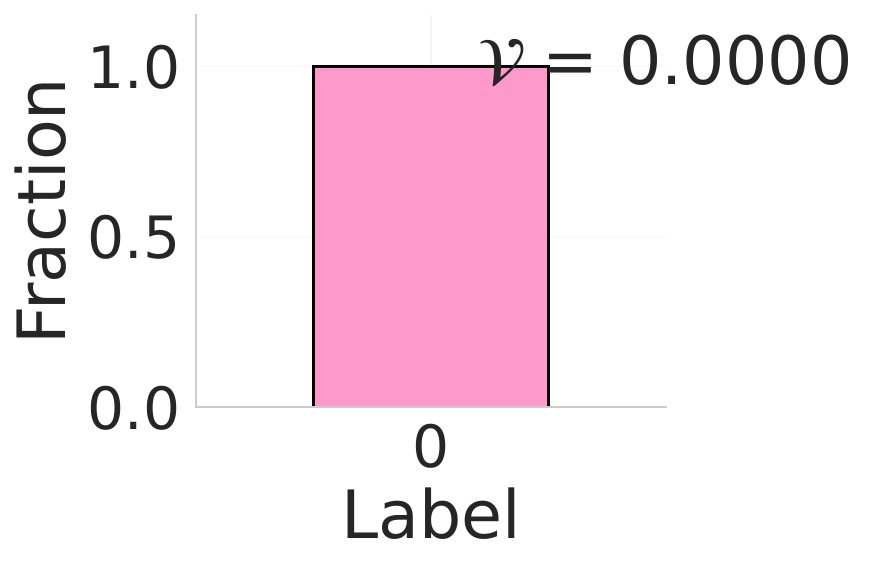}
\end{minipage}
\caption{{\bf Patient-level mortality results on MIMIC-III under missing time-series inputs.} Each panel shows the fraction of clusters assigned to each mortality class, with $\mathcal{V}_{\text{missing}}$ reported as $\mathcal{V}$. The top row shows high-$\mathcal{V}$ examples with clusters spread across multiple risk classes, while the bottom row shows low-$\mathcal{V}$ examples dominated by a single class.\label{fig:mortality_results_appendix}}

\end{figure*}

\begin{figure*}[htbp]
\centering
\begin{minipage}[t]{0.32\linewidth}
\centering
\includegraphics[width=\linewidth]{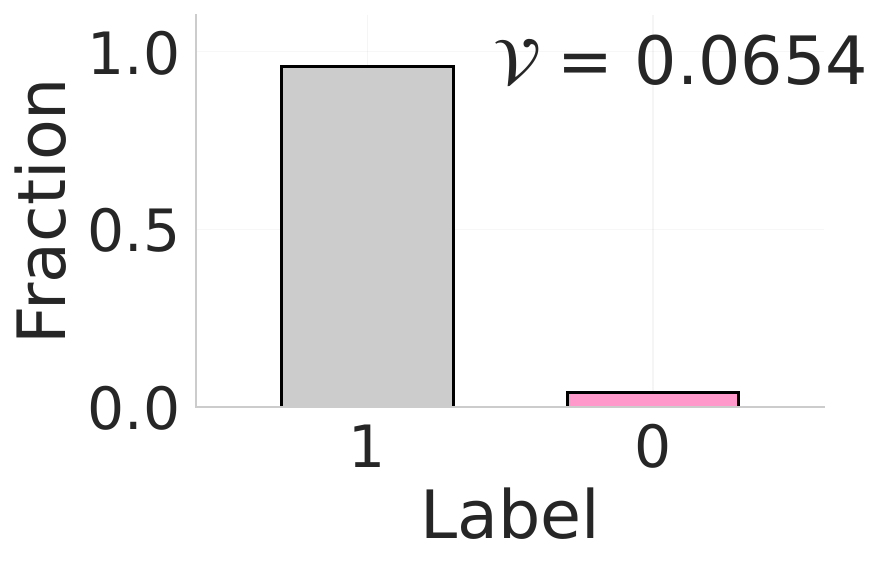}
\end{minipage}\hfill
\begin{minipage}[t]{0.32\linewidth}
\centering
\includegraphics[width=\linewidth]{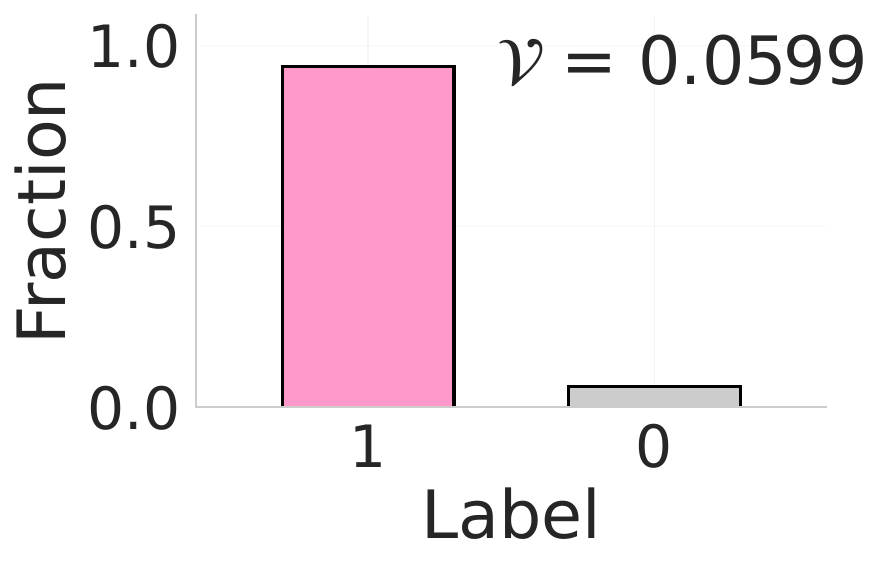}
\end{minipage}\hfill
\begin{minipage}[t]{0.32\linewidth}
\centering
\includegraphics[width=\linewidth]{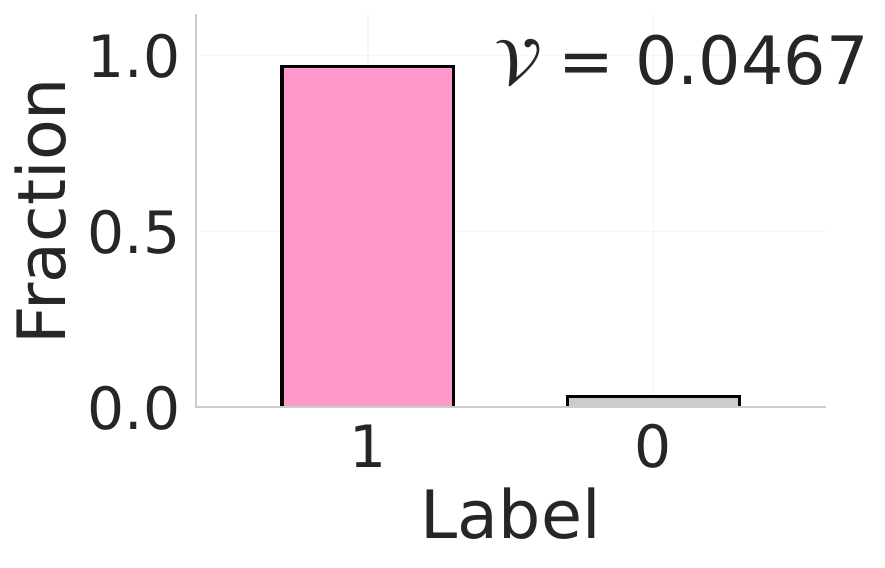}
\end{minipage}

\vspace{1em}

\begin{minipage}[t]{0.32\linewidth}
\centering
\includegraphics[width=\linewidth]{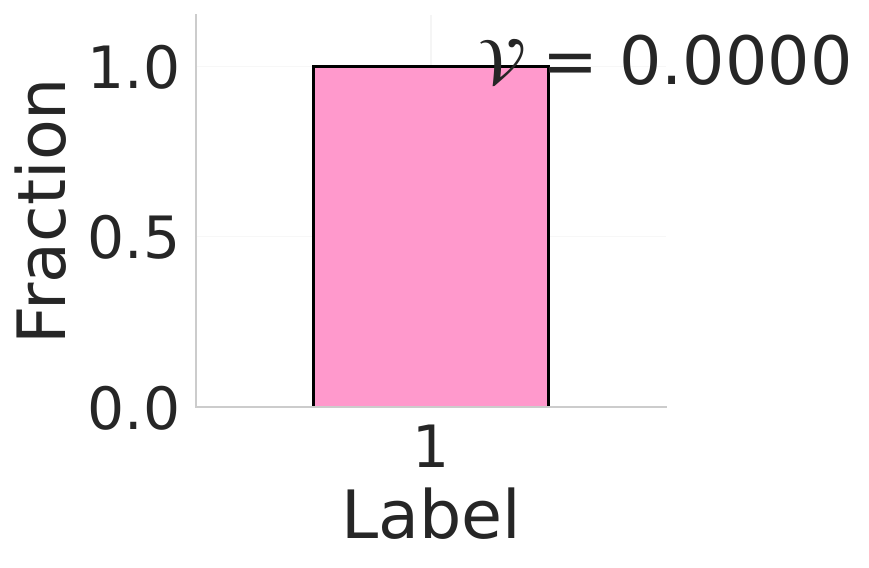}
\end{minipage}\hfill
\begin{minipage}[t]{0.32\linewidth}
\centering
\includegraphics[width=\linewidth]{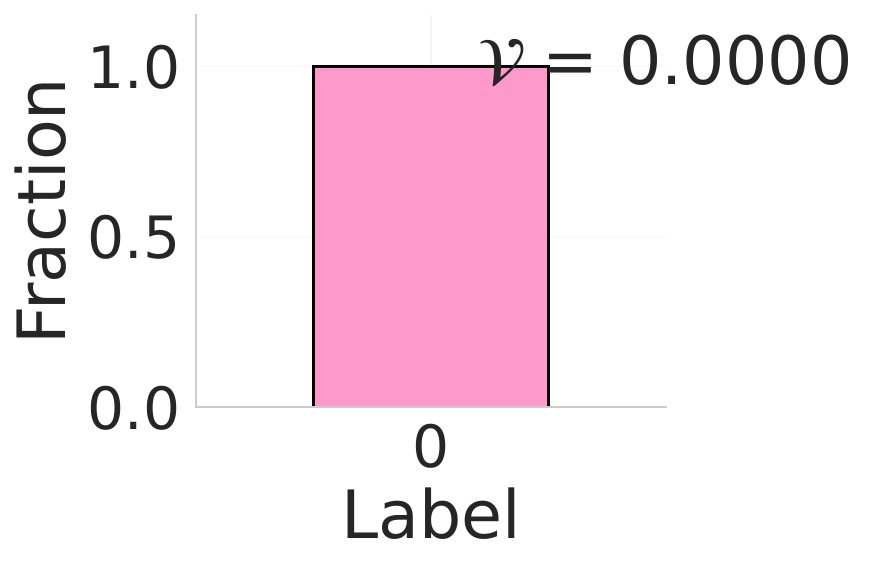}
\end{minipage}\hfill
\begin{minipage}[t]{0.32\linewidth}
\centering
\includegraphics[width=\linewidth]{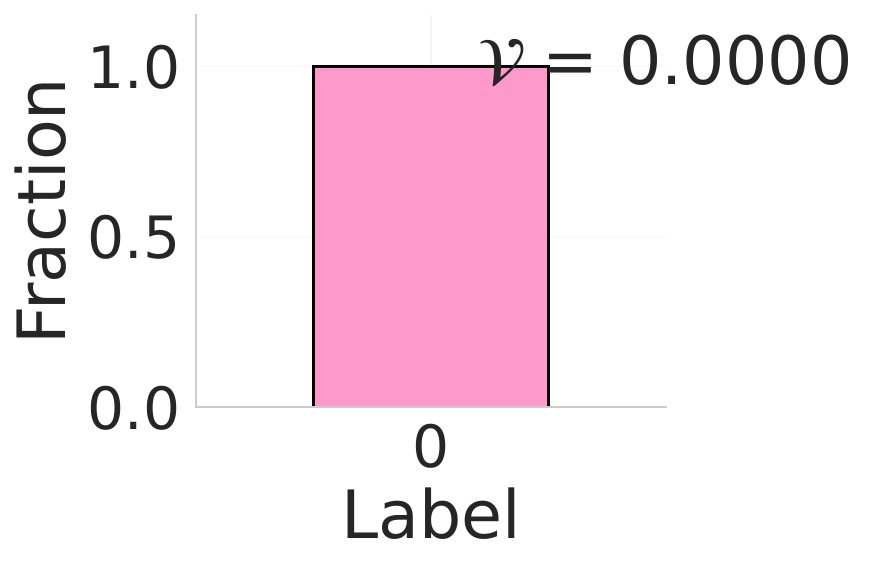}
\end{minipage}
\caption{{\bf Patient-level ICD-9 (140--239) results on MIMIC-III under missing time-series inputs.} Each panel shows the fraction of clusters assigned to each label, with $\mathcal{V}_{\text{missing}}$ reported as $\mathcal{V}$. The top row shows high-$\mathcal{V}$ examples and the bottom row shows low-$\mathcal{V}$ examples; in both cases, clusters are largely dominated by a single label, indicating stable predictions.\label{fig:appendix_icd1_results}}

\end{figure*}

\begin{figure*}[htbp]
\centering
\begin{minipage}[t]{0.32\linewidth}
\centering
\includegraphics[width=\linewidth]{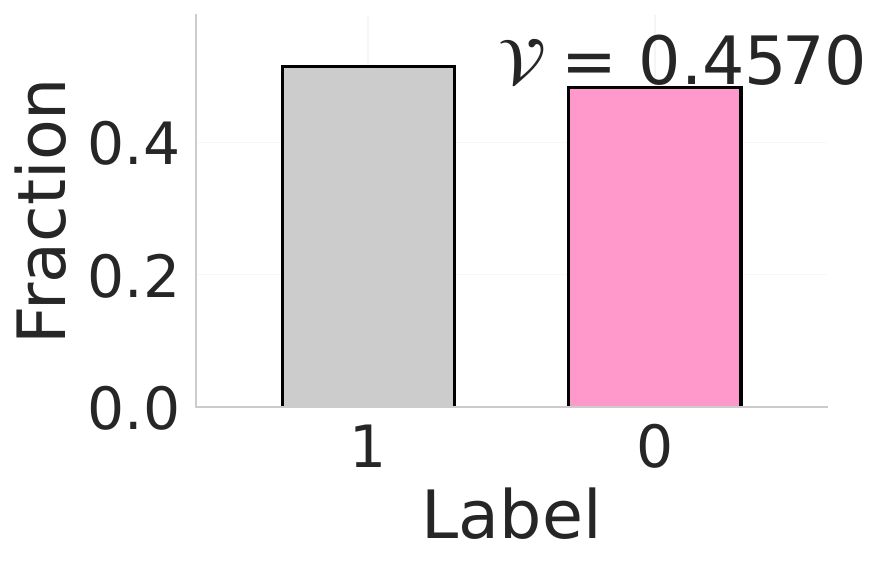}
\end{minipage}\hfill
\begin{minipage}[t]{0.32\linewidth}
\centering
\includegraphics[width=\linewidth]{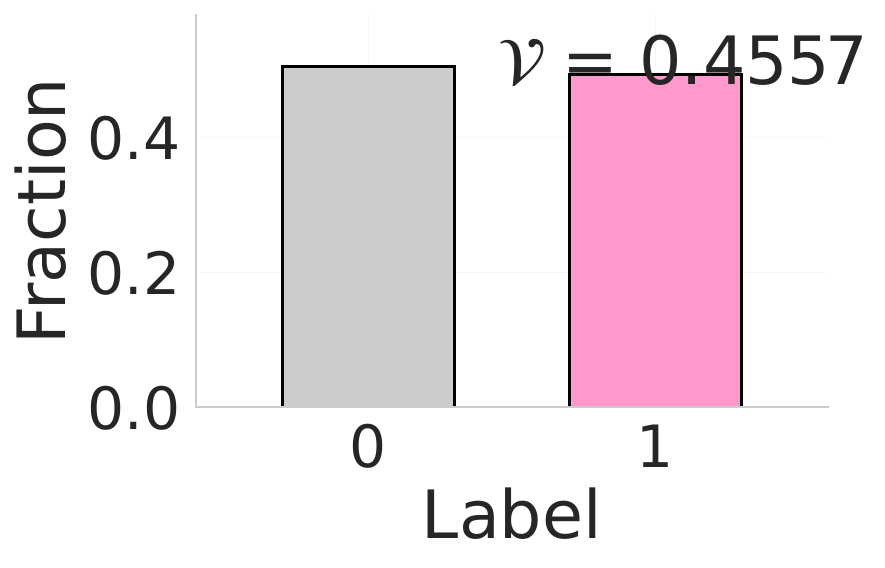}
\end{minipage}\hfill
\begin{minipage}[t]{0.32\linewidth}
\centering
\includegraphics[width=\linewidth]{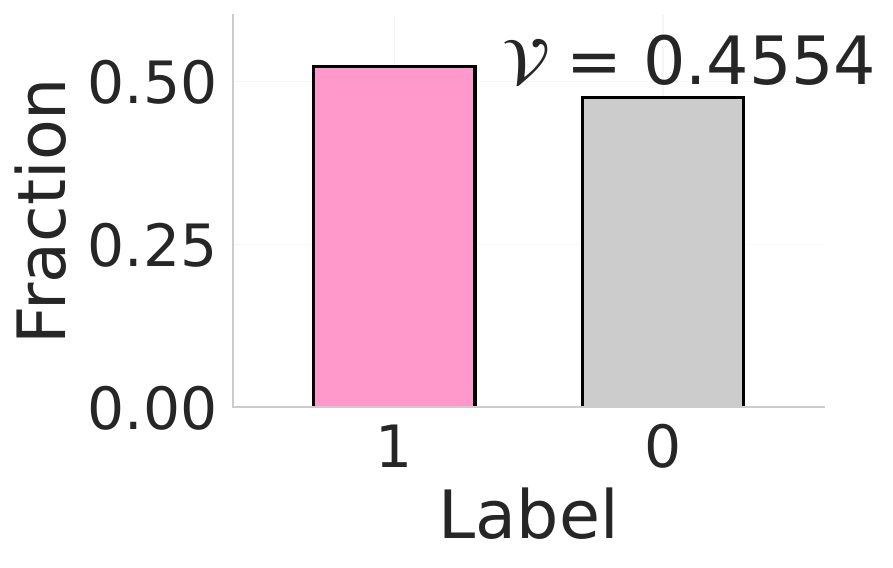}
\end{minipage}

\vspace{1em}

\begin{minipage}[t]{0.32\linewidth}
\centering
\includegraphics[width=\linewidth]{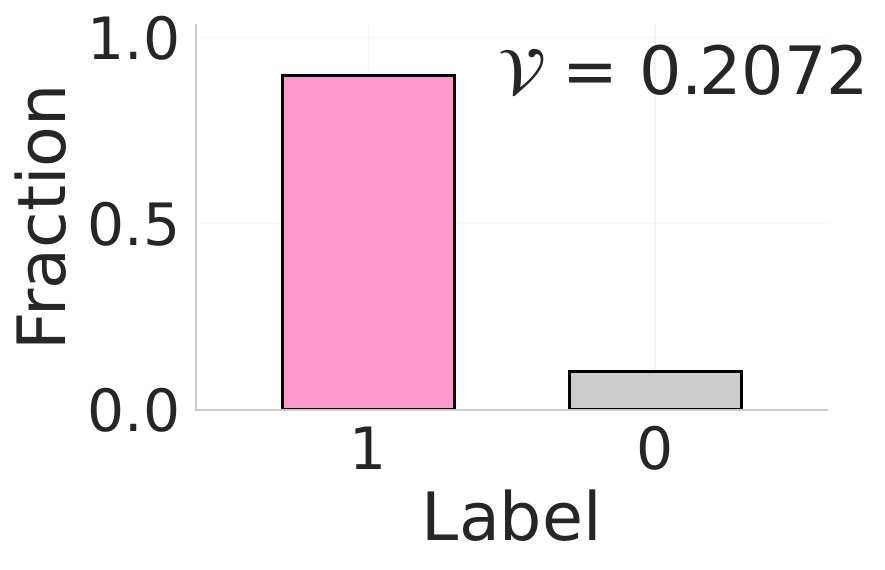}
\end{minipage}\hfill
\begin{minipage}[t]{0.32\linewidth}
\centering
\includegraphics[width=\linewidth]{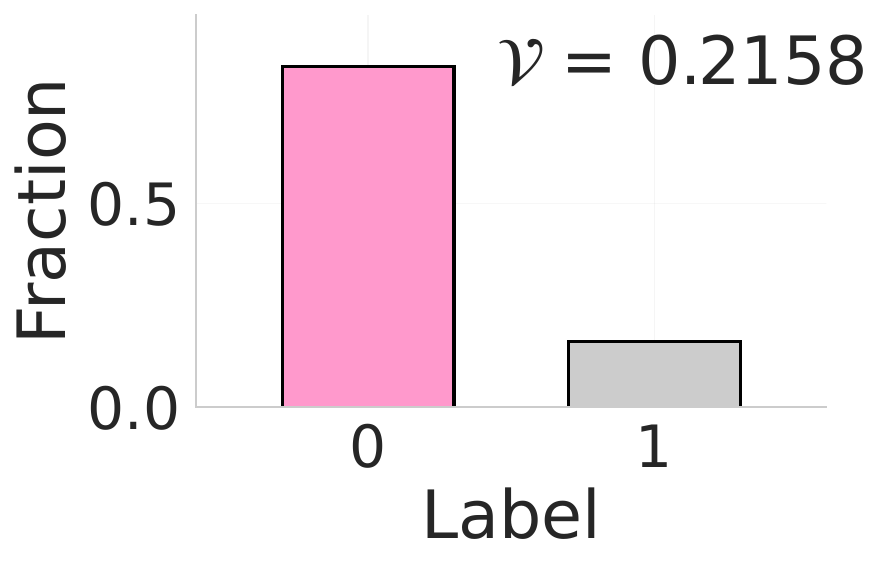}
\end{minipage}\hfill
\begin{minipage}[t]{0.32\linewidth}
\centering
\includegraphics[width=\linewidth]{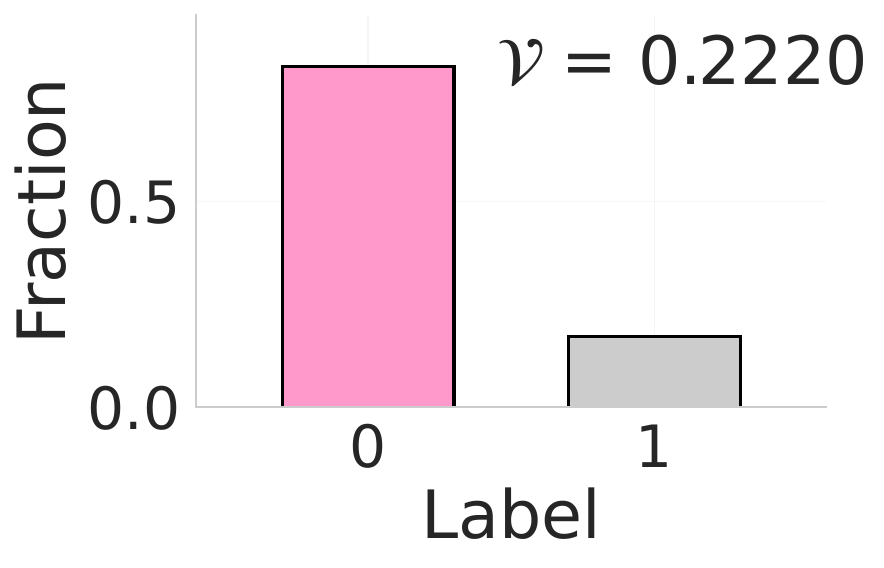}
\end{minipage}
\caption{{\bf Patient-level ICD-9 (460--519) results on MIMIC-III under missing time-series inputs.} Each panel shows the fraction of clusters assigned to each label, with $\mathcal{V}_{\text{missing}}$ reported as $\mathcal{V}$. The top row shows high-$\mathcal{V}$ examples with clusters spread across multiple labels, indicating ambiguity when time-series is missing, while the bottom row shows low-$\mathcal{V}$ examples dominated by a single label.\label{fig:icd2_results}}
\end{figure*}

We report XOR predictions and latent-space visualizations in \Cref{fig:xor_predictions} and \Cref{fig:xor_latent}, respectively. We also provide additional qualitative examples for mortality prediction in \Cref{fig:mortality_results_appendix}, ICD-9 140--239 (neoplasms) prediction in \Cref{fig:appendix_icd1_results}, and ICD-9 460--519 (respiratory diseases) prediction in \Cref{fig:icd2_results}.

\noindent

